\documentclass{article}

\PassOptionsToPackage{numbers, compress}{natbib}
\usepackage[preprint]{neurips_2026}

\usepackage[utf8]{inputenc} 
\usepackage[T1]{fontenc}    
\usepackage{hyperref}       
\usepackage{url}            
\usepackage{booktabs}       
\usepackage{graphicx}       
\usepackage{subcaption}     
\usepackage{amsmath}        
\usepackage{amsfonts}       
\usepackage{nicefrac}       
\usepackage{microtype}      
\usepackage{xcolor}         
\usepackage{cleveref}
\usepackage{wrapfig}
\usepackage{algpseudocode}
\usepackage{needspace}
\usepackage{placeins}
\usepackage{float}
\usepackage{multirow}

\algrenewcommand\algorithmicindent{1.0em}

\newenvironment{compactalgo}[1]{%
  \par\smallskip
  \Needspace{8\baselineskip}
  \noindent\hrule
  \vspace{0.25em}
  \noindent\textbf{#1}
  \vspace{0.25em}
  \hrule
  \vspace{0.25em}
  \scriptsize
  \begin{algorithmic}[1]
}{%
  \end{algorithmic}
  \vspace{-0.25em}
  \hrule
  \smallskip
}

\title{AeroGround: A Comprehensive Benchmark for Aerial-Ground Collaborative Reasoning}

\author{%
    \parbox{0.96\textwidth}{%
            \centering
            Shenghong Yi$^{1,2, \ast}$ \quad
            Lin Zhang$^{2, \ast}$ \quad
            Muzian Li$^{2}$ \quad
            Jiakang Yuan$^{2}$ \quad
            Haoyu Zhang$^{2}$ \\
            Peng Ye$^{2,5}$ \quad
            Jiayuan Fan$^{3}$ \quad
            Huafeng Qin$^{4}$ \quad
            Tao Chen$^{1,2, \dagger}$ \\[0.5em]
            $^{1}$Shanghai Innovation Institute \\[0.15em]
            $^{2}$College of Future Information Technology,
            Fudan University \\[0.15em]
            $^{3}$College of Intelligent Robotics and Advanced Manufacturing,
            Fudan University \\[0.15em]
            $^{4}$Chongqing Technology and Business University \\[0.15em]
            $^{5}$The Chinese University of Hong Kong \\[0.25em]
    }%
}

\begin{document}

\maketitle

\begin{abstract}

Vision-language models (VLMs) have been widely employed in understanding and reasoning tasks for unmanned aerial vehicles (UAVs). Existing UAV benchmarks primarily focus on aerial-view scenarios. However, whether current VLMs can perform well on understanding and reasoning tasks in aerial-ground collaborative scenarios which are practical in real-world applications like rescue and infrastructure inspection remains underexplored. To address this gap, we introduce AeroGround, a comprehensive benchmark for evaluating VLMs in aerial-ground collaborative reasoning. AeroGround is built upon a simulated aerial-ground dataset containing approximately 29,000 multimodal observation groups from diverse open environments, and provides 2,250 high-quality question-answering instances covering cross-view correspondence, spatial understanding, and reasoning. Experiments on 16 pretrained VLMs, together with two domain-adapted variants, reveal a substantial gap between current models and human performance: the best model achieves an average accuracy of 54.4\%, whereas humans reach 93.3\%. By systematically revealing the strengths and limitations of existing models in aerial-ground collaborative reasoning, AeroGround provides a foundation for developing more capable aerial-ground collaborative embodied intelligence systems.
  
\end{abstract}

\section{Introduction}

Unmanned Aerial Vehicles (UAVs) have been widely deployed in open-environment tasks such as search and rescue, disaster assessment, and infrastructure inspection~\cite{jamwal2026comprehensive}. To automate these tasks effectively, UAV AI systems must possess not only basic visual perception capabilities, but also high-level semantic understanding and perceptual reasoning abilities to enable intelligent decision-making and autonomous execution in complex environments. In recent years, Vision-Language Models (VLMs)~\cite{ deepseekvl2_2025,qwen25vl2025}, with their strong visual understanding, language interaction, and cross-modal reasoning capabilities, have been increasingly introduced into UAVs and robotic embodied agents as cognitive models that bridge visual perception and high-level decision-making~\cite{walloss_2025, eo1_2025} . Therefore, establishing systematic benchmarks to assess the spatial understanding capability of VLMs becomes imperative.

Recently, some benchmarks have been proposed to evaluate the understanding and reasoning capability of VLMs under aerial views \cite{dai2025mmuavbench, zhang2025vlmskyready}. However, in real-world scenarios, more complex spatial understanding demands emerge.
Specifically, in tasks such as search and rescue or disaster assessment, UAVs must collaborate with ground personnel or robots to bridge the gap between the aerial global view and the ground local view which provides fine-grained ground-level appearance, close-range interaction states, and local information in occluded regions that are inaccessible to aerial observations.\cite{zacharia2025collaborative} Existing UAV benchmarks typically lack paired aerial-ground heterogeneous views, making it difficult to systematically evaluate models’ abilities in cross-view information integration and unified spatial understanding in aerial-ground collaborative scenarios. On the other hand, existing aerial-ground or cross-view datasets\cite{hou2025agcdrive, wang2026griffin}, although containing both aerial and ground images, are mostly designed for conventional vision tasks such as geo-localization and collaborative target recognition. These datasets typically evaluate visual perception models, have relatively limited task formats, and lack language-guided, multi-task question-answering evaluation designs tailored for VLMs.

\begin{figure}[t]
  \centering
  \includegraphics[width=\linewidth, trim=0 30 0 70, clip]{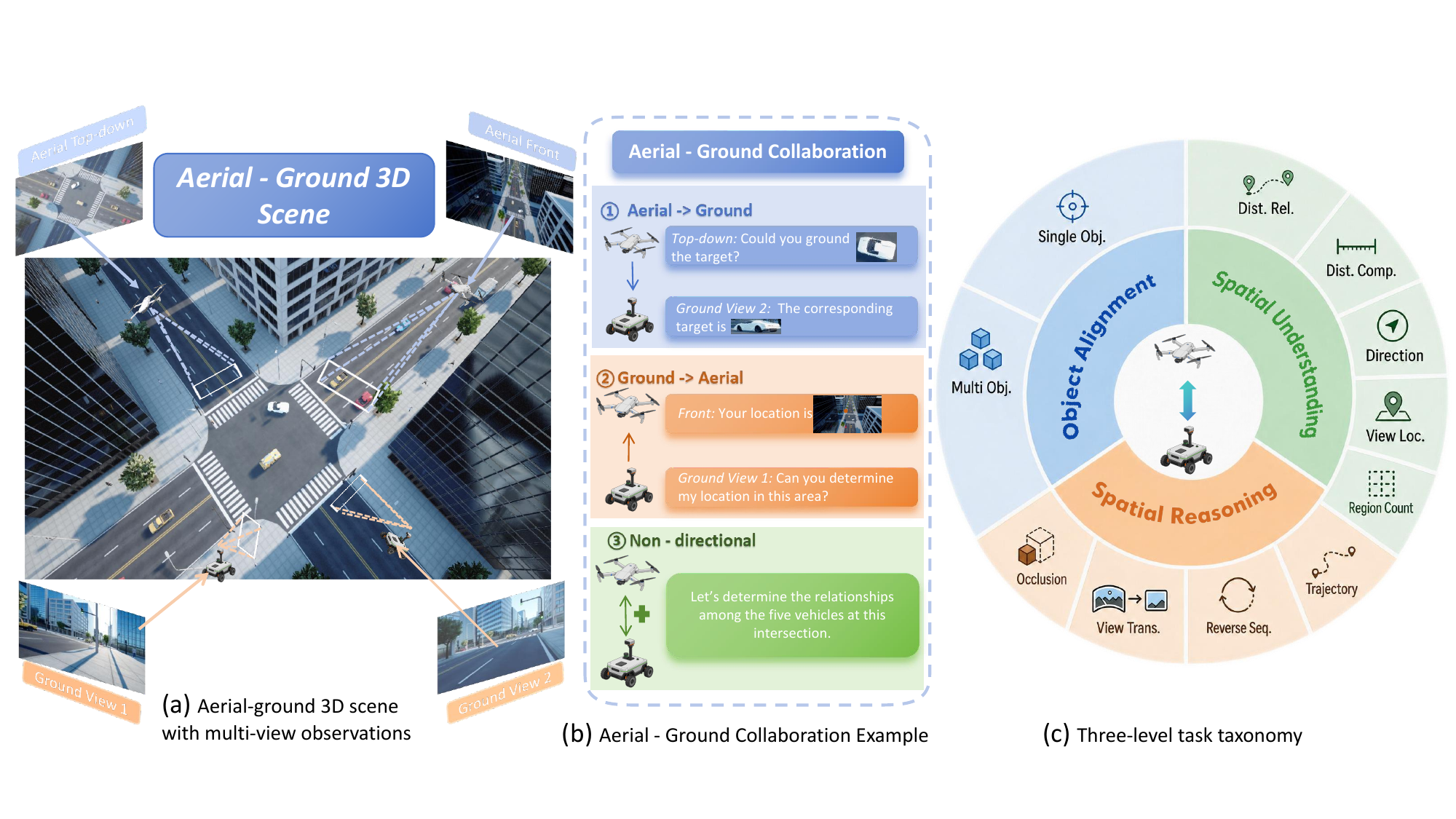}
  \caption{Overview of the aerial-ground dual-view benchmark. 
  (a) Multi-view aerial-ground scene observations. 
  (b) Examples of three directional subject types. 
  (c) Three-level task taxonomy with 11 tasks.}
  \label{fig:benchmark-design}
  \vspace{-0.5em}
\end{figure}

To fill this gap, we introduce AeroGround Dataset and AeroGround Benchmark for systematically evaluating the spatial understanding and reasoning capabilities of VLMs in aerial-ground dual-view scenarios. 
The dataset is constructed in virtual simulation environments \cite{epicgames2021unreal427,shah2018airsim} and contains approximately
29K multimodal observation groups collected from diverse open scenes, including RGB images, depth point clouds, camera poses, and instance segmentation. Based on this dataset, we further construct the benchmark as shown in Figure~\ref{fig:benchmark-design}. It contains 3 capability levels, 11 task categories, and 2,250 high-quality question-answering instances. It further features three different directional subject types, including aerial-to-ground, ground-to-aerial,  and non-directional settings. The benchmark targets heterogeneous spatial understanding between the aerial global view and the ground local view, providing a reproducible testbed for evaluating VLMs’ unified spatial modeling capability in aerial-ground collaborative scenarios.

Comprehensive results on the AeroGround reveal significant limitations in current VLMs' aerial-ground collaborative understanding and reasoning. Doubao-seed-2.0-pro achieves the highest overall accuracy of 54.4\%. However, this performance still lags far behind human performance (93.3\%). Further analysis reveals a complex interplay of factors behind this gap, including weak cross-view correspondence, unstable spatial registration, lacking coordinate transformation, and poor occlusion modeling. We further perform supervised fine-tuning (SFT) on Qwen3-VL-2B/8B-Instruct with the AeroGround Dataset and achieve improvement, showcasing its value for developing more powerful aerial-ground understanding models.

Our contributions are summarized as follows:
\begin{itemize}
  \item We propose the \textbf{AeroGround Dataset}, a large-scale simulated multimodal dataset containing approximately 29K multimodal observation groups from diverse open environments. Our automated dataset construction pipeline provides a scalable foundation for aerial-ground data generation.
  \item We construct the \textbf{AeroGround Benchmark}, the first benchmark for evaluating VLMs in aerial-ground collaborative reasoning. It contains 2,250 question-answer instances covering 3 capability levels, 11 task types, and three complementary evaluation dimensions.
  \item We conduct a comprehensive evaluation of representative VLMs, showing that current models still exhibit a substantial gap from human performance in aerial-ground understanding. Further analyses identify key limitations in cross-view spatial reasoning, while fine-tuning results validate the dataset's value for model adaptation.
\end{itemize}

\section{Related Work}

\subsection{UAV-Oriented Vision-Language Benchmarks}
Recent studies have begun to investigate the spatial understanding and reasoning abilities of VLMs in UAV scenarios. SpatialSky-Bench \citep{zhang2025vlmskyready} evaluates spatial intelligence for UAV navigation, while UAVBench and UAVIT-1M \citep{zhan2026uavbench} construct large-scale low-altitude UAV vision-language tasks and instruction-tuning data. Another line of work further expands the scope of UAV-oriented vision-language evaluation. For example, MM-UAVBench \citep{dai2025mmuavbench} evaluates perception, cognition, and planning in low-altitude UAV scenarios; DVGBench \citep{zhou2026dvgbench} focuses on visual grounding in drone imagery; and UAVReason \citep{sun2026uavreason} builds a unified benchmark for reasoning and generation in nadir-view aerial scenes. These works advance the systematic evaluation of vision-language abilities under UAV viewpoints, but their settings remain largely limited to single aerial views or aerial multi-view inputs, without an explicit aerial-ground collaborative design. In contrast, many real UAV tasks require models to align global aerial context with local ground observations, establish object correspondences across heterogeneous viewpoints, and reason over shared spatial structure. Therefore, as described in Section~\ref{sec:aerial-ground-benchmark}, our benchmark directly evaluates VLMs' unified spatial understanding in aerial-ground collaborative scenarios through paired aerial-ground views and cross-view language-guided reasoning tasks.

\subsection{Aerial-Ground Cooperative Perception}
Another related line of research studies aerial-ground cooperative perception, mainly in autonomous driving and cooperative 3D perception.\cite{han2024collaborative} Earlier V2X datasets, such as OPV2V \citep{xu2022opv2v}, DAIR-V2X \citep{yu2022dairv2x}, and V2X-Real \citep{xiang2024v2xreal}, focus on cooperative perception among vehicles, roadside infrastructure, and other sensor platforms. Recent works such as Griffin \citep{wang2026griffin}, AGC-Drive \citep{hou2025agcdrive}, and V2U4Real \citep{li2026v2u4real} further incorporate UAV observations for aerial-ground cooperative detection, tracking, and 3D perception in driving scenarios. These datasets are valuable for cross-view sensor fusion and cooperative perception, but their task objectives are typically 3D detection, tracking, or point-cloud fusion, primarily serving autonomous-driving perception pipelines. In contrast, the dataset construction and the benchmark design in Section~\ref{sec:aerial-ground-benchmark} focus on VLM-oriented aerial-ground collaborative understanding and reasoning, aiming to evaluate whether models can understand paired heterogeneous aerial-ground views through natural-language tasks and reason over shared spatial structure.

\section{AeroGround Benchmark}
\label{sec:aerial-ground-benchmark}

\subsection{Overview}

The AeroGround Benchmark is designed to evaluate the ability of vision-language models to understand and reason over paired aerial-ground observations in collaborative embodied scenarios. The benchmark is built upon the AeroGround Dataset, which is collected from simulated open environments and contains approximately 29K multimodal observation groups, covering diverse open-world scenarios with varied spatial layouts, object distributions, and scale variations.

\begin{figure}[H]
  \centering
  \setlength{\abovecaptionskip}{4pt}
  \setlength{\belowcaptionskip}{-4pt}
  \includegraphics[
    width=0.94\linewidth,
    height=0.40\textheight,
    keepaspectratio
  ]{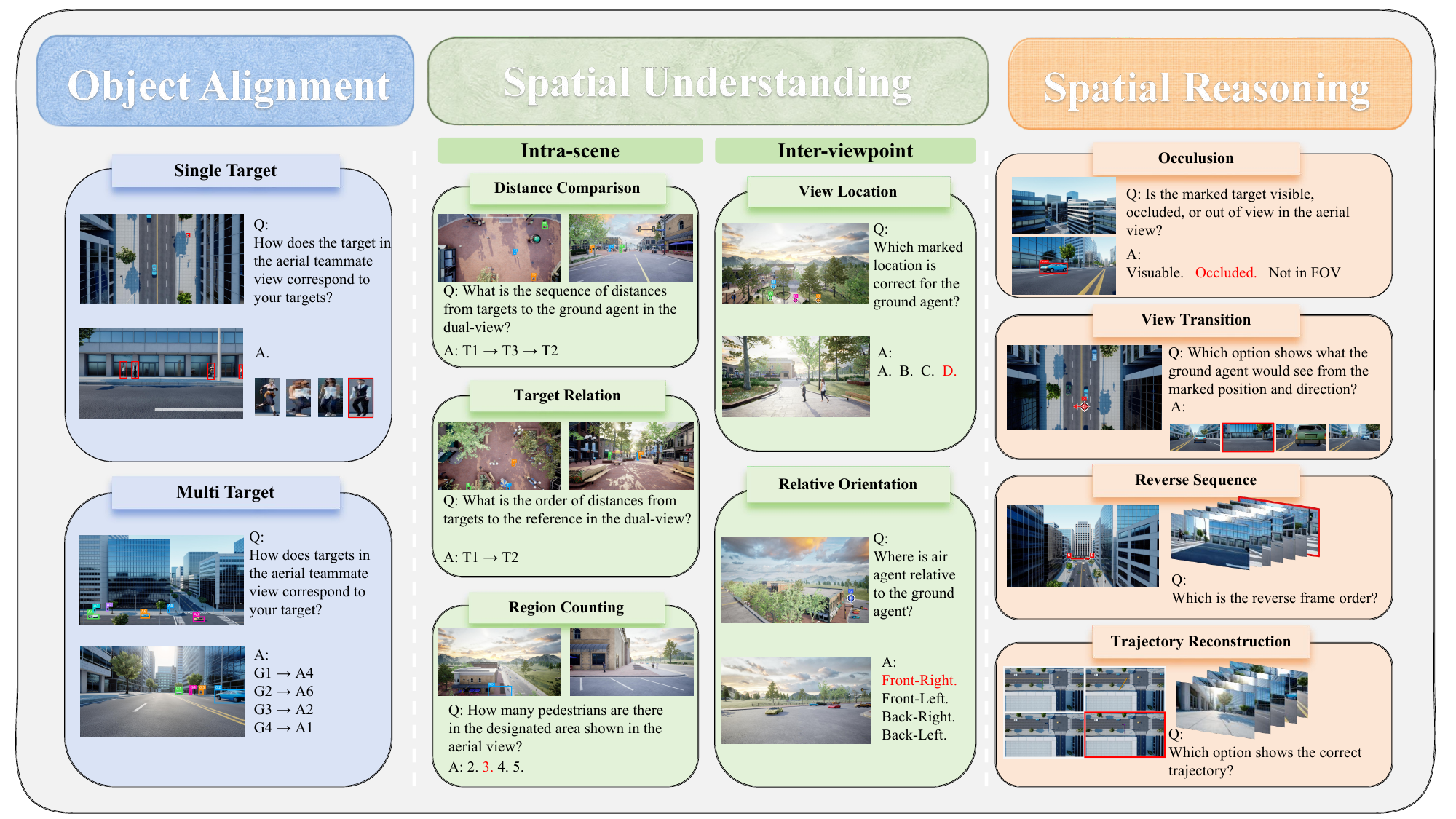}
  \caption{
  Overview of the AeroGround benchmark task taxonomy. AeroGround is organized into three capability groups: object alignment, spatial understanding, and spatial reasoning, covering representative aerial--ground matching, localization, relation, and reasoning tasks.
  }
  \label{fig:benchmark-pipeline}
\end{figure}

Based on this dataset, we construct 2,250 high-quality question-answering instances to systematically evaluate spatial understanding and reasoning from aerial--ground perspectives.\citep{DBLP:journals/corr/abs-1811-00491}\citep{liu2023visualspatialreasoning}\citep{shi2019spatial} As shown in Figure~\ref{fig:benchmark-pipeline}, the benchmark is organized along four complementary dimensions: directional subject, view configuration, spatial span, and task design. The task design dimension comprises 3 capability levels and 11 task types, while the other three dimensions support complementary diagnostic analysis. Through this multi-dimensional design, the benchmark evaluates cross-view understanding and spatial reasoning under different aerial--ground viewpoint conditions, providing a structured testbed for analyzing unified spatial understanding in aerial-ground collaborative scenarios.

\subsection{Dimension Design}

\textbf{Directional Subject.}
Directional subject characterizes the target viewpoint to which an answer is anchored in aerial-ground dual-view tasks, as well as the resulting direction of cross-view evidence transfer. We define three categories: (1) \textbf{Aerial-targeted} (\textit{Ground-to-Aerial}), where the model answers in the aerial view based on ground-view queries; (2) \textbf{Ground-targeted} (\textit{Aerial-to-Ground}), where the model answers in the ground view based on aerial-view queries; and (3) \textbf{Non-directional}, where no dominant cross-view transfer direction is imposed and the model relies on joint view fusion or local spatial cues. This dimension enables analysis of how the answer subject and transfer direction affect model performance and reveals directional biases in aerial-ground  reasoning.

\textbf{View Configuration.}
View configuration describes the geometric relationship between aerial and ground observations, which influences the difficulty of cross-view alignment and unified spatial understanding\citep{8954224}. We distinguish aerial observations into \textbf{top-down} and \textbf{front-view} configurations. For front-view aerial observations, we further consider the relative viewing directions of the aerial and ground subjects, resulting in \textbf{same-side} and \textbf{opposite-side} configurations. This dimension allows us to examine whether models can maintain robust unified spatial understanding under different viewpoint geometries, particularly when aerial and ground cameras observe the same scene from substantially different directions.

\textbf{Spatial Span.}
Spatial span characterizes the scene scale covered by paired aerial-ground observations and is closely related to observation granularity and visual overlap. We describe it using the relative 3D distance between the aerial and ground subjects, which reflects their spatial separation within the scene. This dimension enables analysis of model performance across different aerial-ground spatial ranges, from closely coupled views with strong local correspondence to distant views that require broader spatial reasoning and more robust cross-view alignment.

\textbf{Task Design.}
Task design evaluates aerial-ground collaboration through 3 capability levels and 11 tasks, progressing from basic cross-view association to spatial understanding and higher-level reasoning over a shared scene representation\citep{7410808}\citep{Hu_2018_CVPR}\citep{zheng2020university1652multiviewmultisourcebenchmark}. Specifically, (1) \textbf{Object Alignment} evaluates cross-view object correspondence; (2) \textbf{Spatial Understanding} examines the organization of objects, subjects, and viewpoints in a shared aerial-ground space; and (3) \textbf{Spatial Reasoning} evaluates inference over missing or implicit spatial information under partial observations and multi-view constraints.\citep{zhang2022crossviewimagesequencegeolocalization}

\section{Dataset and Benchmark Processing}
\label{sec:dataset-processing}

\subsection{Data Collection and Pre-processing}

\textbf{Environment Construction.}
We build a unified simulation platform in Unreal Engine by integrating diverse high-fidelity scene assets. The platform covers multiple open-world scenarios, including urban areas, towns, factories, forests, and deserts. To better simulate typical objects in aerial-ground collaboration, we populate these environments with abundant static instances, such as vehicles and pedestrians. This design introduces diverse spatial layouts, scale variations, and occlusion patterns, providing a controllable and realistic foundation for aerial-ground data collection, view alignment, and task generation.

\textbf{Aerial-Ground Data Collection and Annotation.}
Within the constructed environments, we follow a unified automated protocol for aerial-ground data collection. Using the interfaces provided by Unreal Engine 4.27\cite{epicgames2021unreal427} and the AirSim simulation plugin\cite{shah2018airsim}, we collect multiple forms of metadata, including RGB images, depth point clouds, poses, and instance segmentation. We further combine these frame-level observations with global instance information from the simulator to automatically align instances in each view. This process produces frame-level annotations that associate image observations with scene-level object identities and geometric states.

\textbf{Automatic View Pairing.}
Based on frame-level instance alignment, we filter raw observations into paired aerial-ground views. For an aerial frame $a$ and a ground frame $g$, let $\mathcal{U}_a$ and $\mathcal{U}_g$ denote their valid globally aligned instance sets. Their shared evidence and scene center are
\begin{equation}
  \mathcal{S}_{a,g} = \mathcal{U}_a \cap \mathcal{U}_g, \qquad
  \mathbf{c}_{a,g} = \frac{1}{|\mathcal{S}_{a,g}|}\sum_{u \in \mathcal{S}_{a,g}} \mathbf{p}_u ,
  \label{eq:pair-shared-center}
\end{equation}
where $\mathbf{p}_u$ is the simulator-provided 3D position of instance $u$. A pair is retained by three main criteria:
\begin{equation}
  |\mathcal{S}_{a,g}| \geq \tau_s, \qquad
  d_{xy}(a,g) \leq \tau_{xy}, \qquad
  \theta_{a,g} \leq \tau_\theta .
\label{eq:pair-constraints}
\end{equation}
Here $d_{xy}(a,g)$ is the horizontal camera distance, and $\theta_{a,g}$ is the maximum angle between each camera's forward direction and the shared scene center. Among valid candidates, we further favor pairs with stronger shared evidence and more reliable instance alignment. This produces paired aerial-ground observations that share concrete scene evidence while preserving diverse spatial spans and view configurations. In addition, by combining instance alignment with depth point clouds, we automatically compute accurate occlusion states of target objects\cite{richter2016playing}. We also leverage information from adjacent frames to pair view sequences and construct trajectory-level samples. Through these steps, raw simulated observations are converted into structured sample resources suitable for evaluating aerial-ground collaborative understanding.

\subsection{Question-Answer Generation}

\textbf{Multi-Task Question-Answer Generation.}
After aerial-ground view pairing, we use the collected and annotated metadata to generate question-answering instances. For each task type, we design task-specific generation rules that encode the required spatial and semantic constraints.\citep{DBLP:journals/corr/abs-1902-09506} These rules ensure that each question has an unambiguous answer, which can be verified using simulator-derived ground-truth data rather than relying solely on human interpretation\cite{johnson2017clevr, chen2024spatialvlm}.

\textbf{Human Filtering and Calibration.}
To improve the reliability of the benchmark, we manually review and filter the automatically generated question-answering instances. Samples with ambiguous wording, unstable view pairing, unreliable annotations, or unclear visual evidence are removed. After this quality-control process, the final benchmark contains 2,250 high-quality question-answering instances, covering 3 capability levels and 11 tasks. This process balances the scalability of automatic generation with human quality control, making the final evaluation more reliable and interpretable.
\section{Experiment}

\begin{table}[!t]
  \caption{Main benchmark results. Entries are accuracies (\%); higher is better. Avg. is computed over all questions. Bold and underline indicate the best and second-best task scores within each model group, respectively.}
  \label{tab:main-results}
  \centering
  \scriptsize
  \setlength{\tabcolsep}{2.4pt}
  \resizebox{\textwidth}{!}{%
  \begin{tabular}{lcccccccccccc}
    \toprule
    & & \multicolumn{2}{c}{Obj. Align.} & \multicolumn{5}{c}{Spatial Underst.} & \multicolumn{4}{c}{Spatial Reason.} \\
    \cmidrule(lr){3-4} \cmidrule(lr){5-9} \cmidrule(lr){10-13}
    Model & Avg. & Single & Multi & Rel. & Dist. & Count & Ori. & Loc. & Occ. & ViewTr. & RevSeq. & Traj. \\
    \midrule
    \multicolumn{13}{l}{\textit{Reference}} \\
    Random Baseline & 24.6 & 23.6 & 0.0 & 35.0 & 23.2 & 15.6 & 23.3 & 22.7 & 34.4 & 32.8 & 22.5 & 23.0 \\
    Human Level & 93.3 & 100.0 & 93.3 & 83.3 & 93.3 & 93.3 & 96.7 & 90.0 & 86.7 & 96.7 & 96.7 & 96.7 \\
    \midrule
    \multicolumn{13}{l}{\textit{Closed-source models}} \\
    GPT-5.4 & 40.1 & \underline{37.4} & 14.7 & 68.9 & 48.8 & \textbf{39.1} & \underline{30.0} & 38.0 & \underline{42.2} & 33.9 & \underline{41.2} & \underline{26.0} \\
    Gemini-3.1-Pro & \underline{47.8} & 37.0 & \underline{25.9} & \underline{78.9} & \textbf{71.2} & \underline{23.4} & \underline{30.0} & \textbf{48.7} & 36.6 & \textbf{71.0} & 35.0 & 21.0 \\
    Doubao-seed-2.0-pro & \textbf{54.4} & \textbf{50.0} & \textbf{30.6} & \textbf{83.3} & \underline{69.0} & 20.3 & \textbf{43.3} & \underline{45.3} & \textbf{45.0} & \underline{64.0} & \textbf{77.5} & \textbf{45.0} \\
    \midrule
    \multicolumn{13}{l}{\textit{Open-source models}} \\
    Kimi-K2.5 & \underline{39.3} & \textbf{39.8} & \textbf{17.6} & 53.3 & \underline{58.5} & 37.5 & 11.1 & 34.0 & 38.1 & 30.1 & 5.0 & \textbf{53.0} \\
    Qwen3-VL-8B-Thinking & 28.8 & 28.0 & 10.0 & 52.8 & 36.8 & \underline{39.1} & 7.8 & 16.7 & 28.4 & 36.0 & 11.2 & 21.0 \\
    InternVL3.5-8B & 31.0 & 35.2 & 8.8 & 45.0 & 37.3 & 35.9 & \underline{27.8} & 18.7 & 34.1 & 25.3 & 13.8 & 29.0 \\
    GLM-4.1V-9B-Base & 34.5 & 32.4 & 5.3 & 41.7 & 48.0 & 37.5 & \textbf{34.4} & 32.7 & \underline{40.3} & 28.5 & 17.5 & \underline{33.0} \\
    LLaVA-OneVision-7B & 30.5 & 27.4 & 1.2 & 50.0 & 34.1 & 34.4 & 26.7 & 30.7 & 36.6 & 38.7 & 15.0 & 24.0 \\
    Qwen3-VL-32B-Instruct & 37.4 & 33.6 & 8.8 & 51.7 & 50.7 & 34.4 & 18.9 & \textbf{38.0} & 38.8 & 38.2 & \textbf{50.0} & 27.0 \\
    Qwen3-VL-32B-Thinking & \textbf{41.6} & 37.2 & \underline{14.1} & \textbf{68.9} & \textbf{62.7} & \textbf{45.3} & 11.1 & \underline{34.7} & 37.5 & \underline{40.3} & \underline{42.5} & 26.0 \\
    InternVL3.5-38B & 38.8 & \underline{38.6} & 10.0 & \underline{62.8} & 50.0 & 35.9 & 16.7 & 24.7 & \textbf{42.8} & \textbf{48.4} & 27.5 & 22.0 \\
    \midrule
    \multicolumn{13}{l}{\textit{Spatial reasoning models}} \\
    Spatial-SSRL-Qwen3VL-4B & \textbf{33.8} & \textbf{35.4} & \textbf{7.6} & \textbf{60.6} & \underline{37.3} & 17.2 & \underline{28.9} & \underline{22.7} & \textbf{36.6} & \textbf{33.3} & \textbf{43.8} & \underline{23.0} \\
    SpaceOm-4B & \underline{30.1} & 28.0 & \underline{0.0} & \underline{33.3} & \textbf{45.6} & \underline{28.1} & \textbf{34.4} & \textbf{31.3} & \textbf{36.6} & 21.5 & 10.0 & \textbf{29.0} \\
    SpaceThinker-Qwen2.5VL-3B & 26.6 & \underline{29.4} & \underline{0.0} & 26.1 & 33.9 & \textbf{34.4} & \underline{28.9} & 18.0 & \underline{32.8} & \underline{26.9} & \underline{17.5} & 21.0 \\
    \midrule
    \multicolumn{13}{l}{\textit{Domain Fine-tuned \& Baselines}} \\
    Qwen3-VL-2B-Instruct & 29.4 & 30.8 & 2.9 & \textbf{51.1} & 34.6 & 21.9 & 23.3 & 24.7 & 34.1 & 28.0 & 23.8 & 17.0 \\
    Qwen3-VL-2B-Instruct-LoRA & 30.9 & \textbf{34.2} & 0.0 & 43.3 & \underline{40.2} & 15.6 & \underline{25.6} & \underline{29.3} & 35.0 & \underline{33.9} & 16.2 & 17.0 \\
    Qwen3-VL-8B-Instruct & \underline{33.3} & \underline{32.0} & \textbf{10.0} & \underline{49.4} & 38.1 & \underline{34.4} & \textbf{27.8} & 24.0 & \textbf{42.8} & 32.8 & \textbf{35.0} & \underline{18.0} \\
    Qwen3-VL-8B-Instruct-LoRA & \textbf{35.5} & \textbf{34.2} & \underline{7.7} & 46.7 & \textbf{42.0} & \textbf{45.3} & 24.4 & \textbf{40.0} & \underline{38.1} & \textbf{39.8} & \underline{33.8} & \textbf{25.0} \\
    \bottomrule
  \end{tabular}}
\end{table}

\subsection{Experimental Setup}

\textbf{Model Evaluation Setup.}
We evaluate 16 pretrained VLMs and additionally report two
domain-adapted variants\cite{openai2026gpt5, deepmind2026gemini31pro, bytedance2024doubao, kimiteam2026kimik25, bai2025qwen3vl, wang2025internvl35, glmvteam2025glm41v, li2024llavaonevision, liu2025spatialssrl, batra2025spatialthinker}, including closed-source models, open-source models, and models with enhanced spatial reasoning capabilities, to comprehensively assess the ability of large multimodal models in aerial-ground collaborative scenarios. All models are evaluated under the same prompt template and image input conditions to ensure comparability. Each task is evaluated using an automatic closed-form protocol, including multiple-choice questions and format matching.

\textbf{Random and Human Baseline Setup.}
For each task type, we construct a probability-based random-choice baseline to estimate chance-level performance under the corresponding answer space. Furthermore, we assign four human evaluators to complete all questions across the 11 tasks, yielding a human baseline for the entire benchmark.

\textbf{Model Training Setup.}
To validate the practical value of the dataset for domain knowledge alignment, we conduct controlled low-rank adaptation\cite{hu2022lora} fine-tuning experiments based on Qwen3-VL-2B and Qwen3-VL-8B. The training set contains 9,475 high-quality samples and adopts a scene-disjoint split from the benchmark test set, where the two sets are constructed from different scene assets. Detailed experimental settings are provided in the appendix.

\subsection{Main results}

\textbf{Model performance remains limited.}
Table~\ref{tab:main-results} shows that the best closed-source model, Doubao-seed-2.0-pro, achieves an overall accuracy of 54.4\% (1225/2250), while the best open-source model, Qwen3-VL-32B-Thinking, reaches 41.6\% (937/2250), leaving a gap of 12.8 percentage points. Although all models substantially outperform the random baseline of 24.6\%, their absolute performance remains insufficient for reliable aerial-ground understanding and reasoning. This indicates that neither closed-source nor open-source models have demonstrated mature unified spatial understanding from aerial-ground perspectives.

\textbf{Task difficulty varies considerably.}
Overall, models perform better on spatial understanding tasks than on object alignment and spatial reasoning tasks. This advantage mainly comes from tasks that rely on local spatial relations or object relations within a single image. However, model performance drops notably when tasks require precise aerial-ground view alignment, coordinate transformation, and occlusion judgment. These results suggest that current models can exploit certain local spatial cues, but still lack stable spatial representations for aerial-ground perspectives.

\textbf{Spatially enhanced models show limited advantages.}
Spatial-SSRL-Qwen3VL-4B still achieves lower overall accuracy than stronger general-purpose vision-language models. This suggests that existing spatial reasoning enhancement methods do not fully address the challenges introduced by heterogeneous aerial and ground views. Therefore, this benchmark not only evaluates general spatial description ability, but also reveals deeper limitations in aerial-ground view alignment, spatial understanding, and reasoning.

\textbf{The human--model gap remains substantial.}
Human evaluators achieve an average accuracy of 93.3\%, significantly outperforming all current models. As shown in Figure~\ref{fig:human-doubao-scatter}, this gap is not uniform across tasks: tasks that are relatively difficult for humans are not necessarily the most difficult for models, while some tasks that humans can solve reliably remain challenging for VLMs. This indicates clear differences between current models and humans in aerial-ground spatial understanding mechanisms.

\textbf{Domain fine-tuning brings effective improvements.}
Controlled low-rank adaptation fine-tuning experiments based on Qwen3-VL show that the constructed dataset provides effective aerial-ground domain knowledge for model adaptation. After fine-tuning, Qwen3-VL-2B and 8B improve their overall accuracy by 1.55\% and 2.22\%, respectively, with the 8B model obtaining a larger performance gain. These results demonstrate that domain alignment training based on our dataset can effectively enhance spatial understanding in aerial-ground scenarios and further validate the dataset's practical value for developing aerial-ground spatial reasoning models.

\subsection{Directional Subject Analysis}

\begin{figure}[t]
  \centering
  \begin{minipage}{0.49\linewidth}
    \centering
    \includegraphics[
      width=\linewidth,
      trim=5 5 5 10,
      clip
    ]{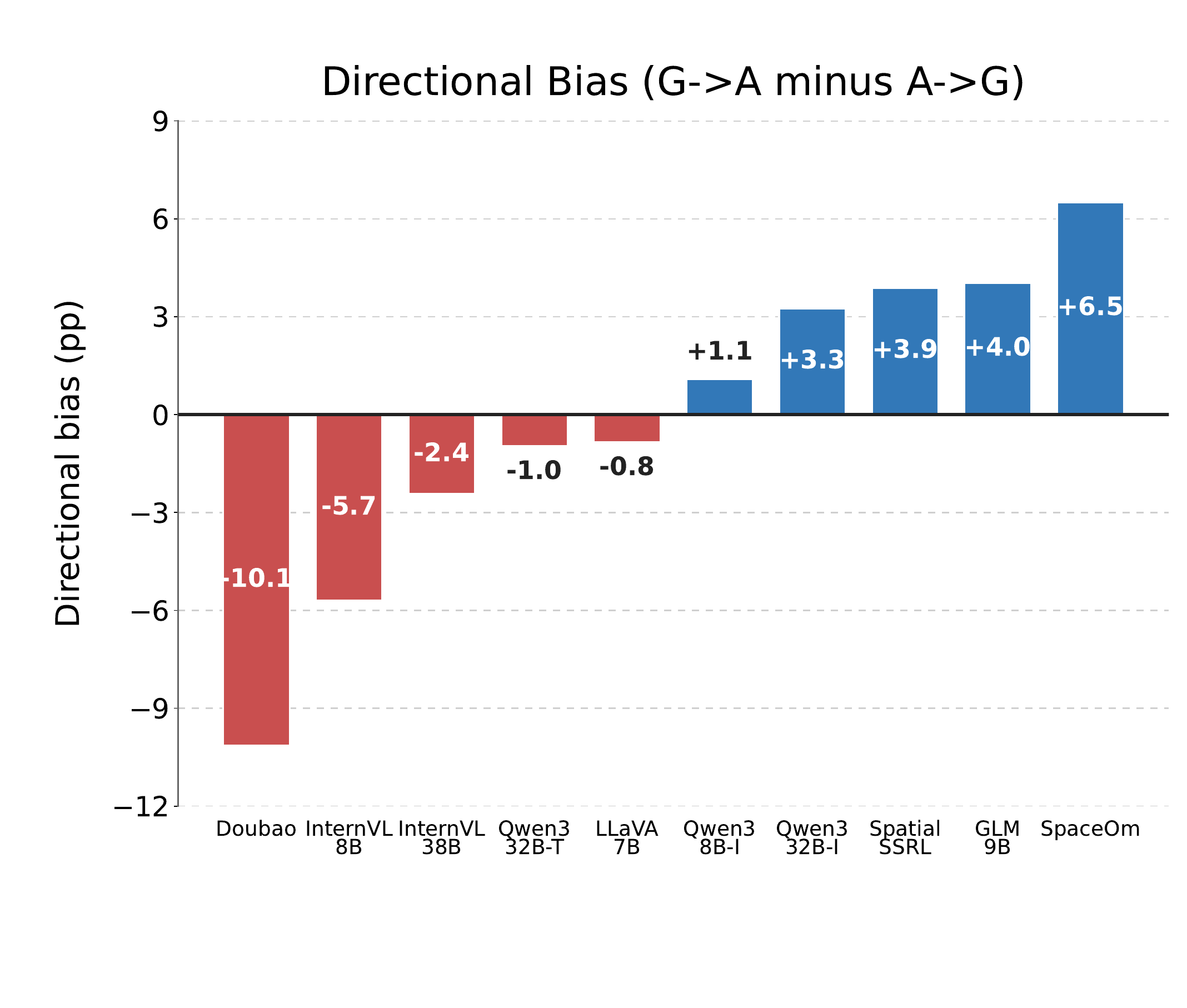}
    \vspace{-0.2em}
    \centerline{\small (a) Directional bias.}
  \end{minipage}
  \hfill
  \begin{minipage}{0.49\linewidth}
    \centering
    \includegraphics[
      width=\linewidth,
      trim=5 5 5 10,
      clip
    ]{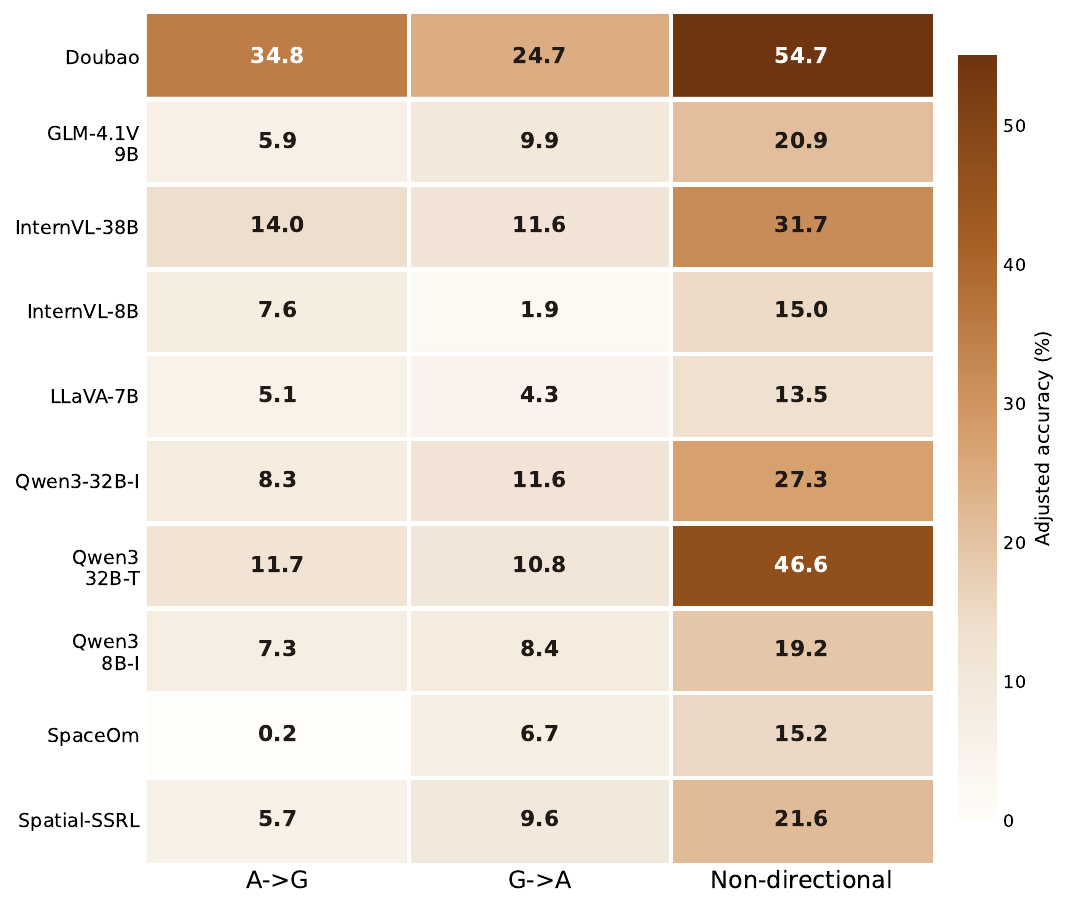}
    \vspace{-0.2em}
    \centerline{\small (b) Group accuracy.}
  \end{minipage}
  \vspace{-0.3em}
  \caption{Directional subject analysis. 
  (a) Chance-adjusted bias between Ground-to-Aerial and Aerial-to-Ground tasks. 
  (b) Chance-adjusted accuracy across directional subject groups.}
  \label{fig:directional-subject-analysis}
  \vspace{-0.5em}
\end{figure}

To avoid confounding comparisons caused by different random baselines across tasks, we analyze the directional subject dimension using chance-adjusted accuracy in Figure~\ref{fig:directional-subject-analysis}. We aggregate Aerial-to-Ground, Ground-to-Aerial, and Non-directional tasks separately to examine how the direction of cross-view transfer affects model performance.

\textbf{Models perform better on non-directional tasks.}
After adjusting for random baselines, non-directional tasks still clearly outperform directional tasks. For the strongest model, Doubao-seed-2.0-pro achieves a chance-adjusted accuracy of 54.7\% on non-directional tasks, which drops to 29.6\% on directional tasks. This trend also holds across all models: the average chance-adjusted accuracy is 24.7\% on non-directional tasks, compared with only 9.2\% on directional tasks. This suggests that models can more effectively exploit joint-view information or intra-view local spatial cues when explicit cross-view evidence transfer is not required, whereas directional tasks introduce additional challenges in aerial-ground information transfer, cross-view registration, and unified spatial modeling.

\textbf{There is no consistent one-way advantage.}
Between Aerial-to-Ground and Ground-to-Aerial directional tasks, models do not show a consistent one-way advantage. Overall, the average chance-adjusted accuracies across all models are close for the two task types, at 9.6\% and 8.8\%, respectively. However, different models exhibit different directional preferences: Doubao-seed-2.0-pro performs better on Aerial-to-Ground tasks, reaching 34.8\%, compared with 24.7\% on Ground-to-Aerial tasks; in contrast, SpaceOm shows the opposite trend, achieving 6.7\% on Ground-to-Aerial tasks and 0.2\% on Aerial-to-Ground tasks. These results indicate that aerial-ground reasoning is challenging in both directions: the former requires grounding global layout cues into local observations, while the latter requires recovering global spatial context from local evidence. Therefore, this dimension not only evaluates overall spatial reasoning ability, but also diagnoses model-specific weaknesses under different directions of cross-view information flow.

\subsection{View configuration analysis}

\setlength{\intextsep}{0.3em}
\begin{wrapfigure}{r}{0.45\linewidth}
  \vspace{-1.0em}
  \centering
  \includegraphics[
    width=\linewidth,
    trim=50 25 25 40,
    clip
  ]{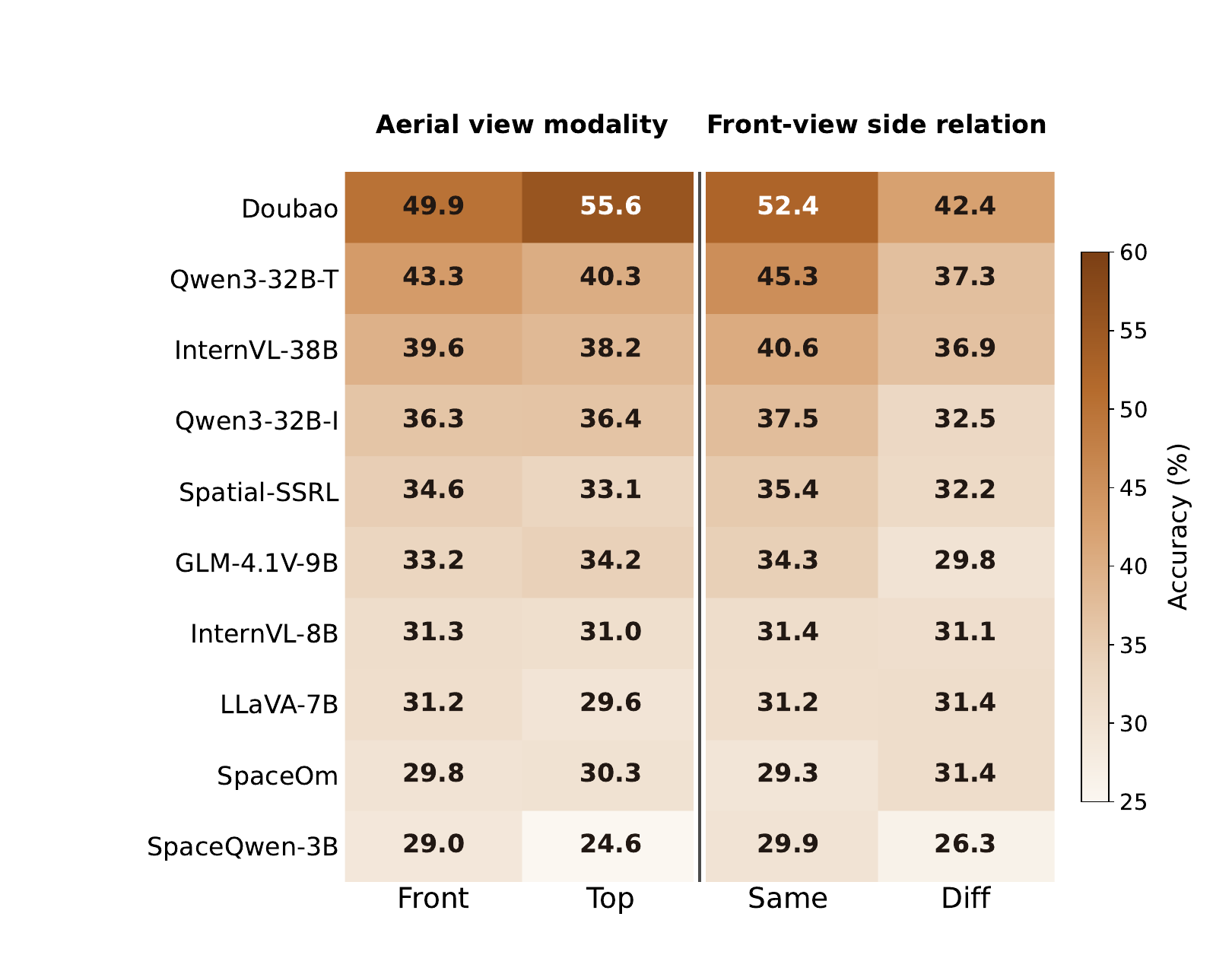}
  \vspace{-0.8em}
  \caption{View configuration analysis. Accuracy across aerial-view modalities and front-view side relations.}
  \label{fig:view-configuration-analysis}
  \vspace{-0.2em}
\end{wrapfigure}

To analyze the effect of view configuration on model performance, we report the overall results under top-down and front-view aerial observations, and further compare same-side and opposite-side configurations within the front-view setting in Figure~\ref{fig:view-configuration-analysis}.

\textbf{Top-down views do not necessarily lead to better performance.}
Results along the view configuration dimension show that top-down aerial observations do not consistently outperform front-view observations. Overall, the average accuracy across all models is slightly higher under front-view settings than under top-down settings, at 35.8\% and 35.3\%, respectively. Although the strongest model, Doubao-seed-2.0-pro, benefits from top-down observations, improving from 49.9\% under front-view settings to 55.6\% under top-down settings, some models show the opposite trend or only minor differences. This suggests that the global layout information provided by top-down views is effective only when the model can reliably align it with the ground-view reference frame; otherwise, the larger visual gap between top-down and ground views may offset its global-structure advantage.

\textbf{Same-side front-view configurations are generally easier.}
Under front-view observations, same-side configurations are generally easier than opposite-side configurations. The average accuracy across all models is 36.7\% under same-side settings, higher than 33.1\% under opposite-side settings. Most models experience performance drops in opposite-side configurations; for example, Doubao-seed-2.0-pro decreases from 52.4\% to 42.4\%, and Qwen3-VL-32B-Thinking decreases from 45.3\% to 37.3\%. This indicates that larger changes in viewing direction make it more difficult for models to maintain consistent spatial relations across aerial and ground views.

\subsection{VLMs Error Analysis}

\begin{figure}[t]
  \centering
  \includegraphics[
    width=0.98\linewidth,
    trim=0 0 0 10,
    clip
  ]{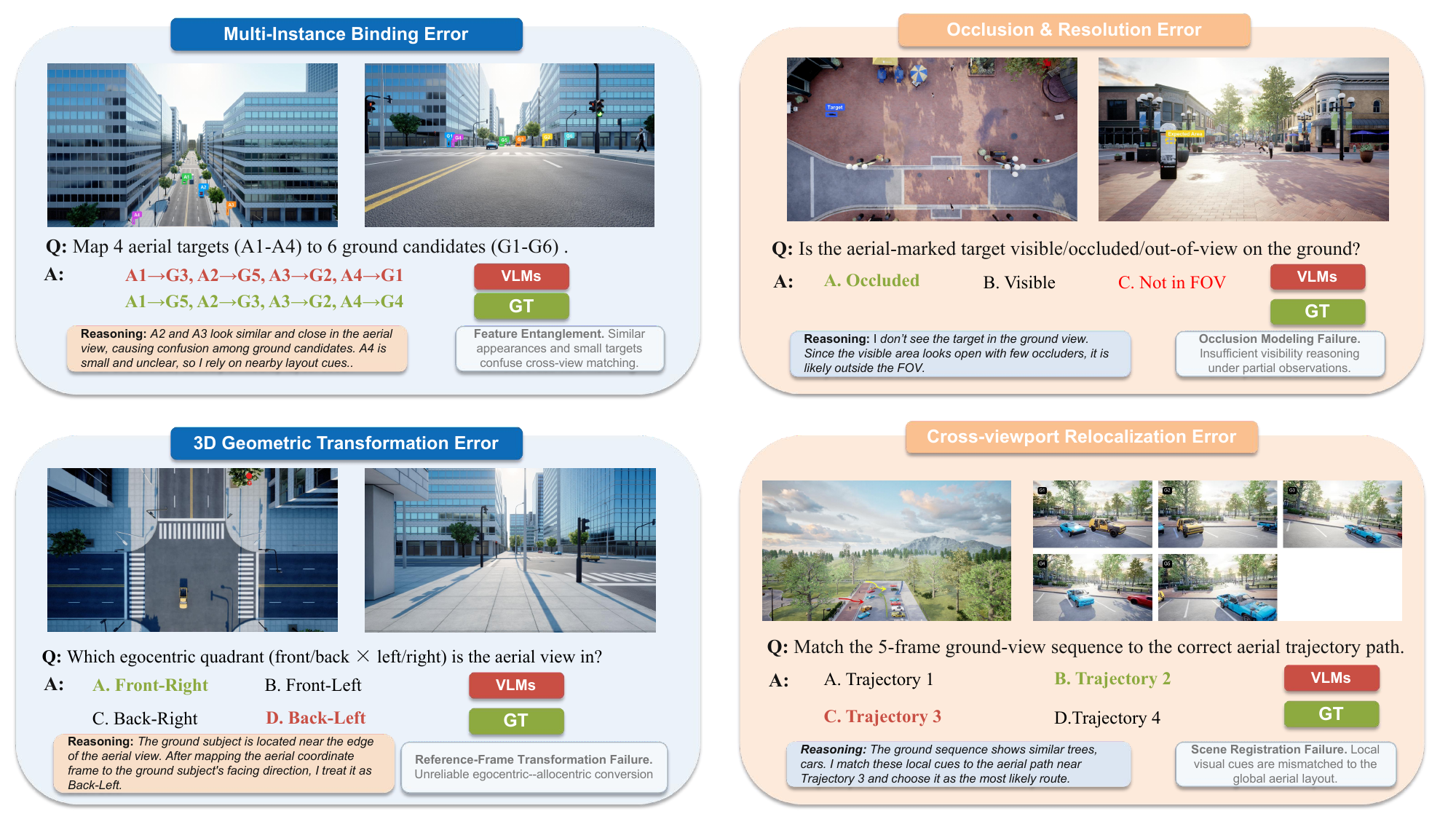}
  \caption{Representative failure cases in aerial-ground dual-view reasoning, covering cross-view correspondence, visibility reasoning, reference-frame transformation, and scene registration.}
  \label{fig:error-analysis}
  \vspace{-1em}
  
\end{figure}

Based on the quantitative results and representative failure cases across tasks, we further summarize the major failure modes of current VLMs in aerial-ground dual-view reasoning in Figure~\ref{fig:error-analysis}. These errors not only reflect task-specific limitations, but also reveal systematic bottlenecks in cross-view correspondence, scene registration, reference-frame transformation, and visibility reasoning.

\textbf{Insufficient cross-view correspondence for small objects.}
Targets in aerial images are often small in scale and densely distributed, while the same targets may exhibit substantially different appearances, scales, and occlusion states in ground images. As shown in Figure~\ref{fig:error-analysis}, this makes it difficult for models to establish stable object-level correspondences across viewpoints. When visual evidence is insufficient, human evaluators can more effectively leverage surrounding context and inter-instance spatial relations to support cross-view matching. Current models remain limited in using contextual cues to compensate for local visual uncertainty.

\textbf{Unstable registration between global aerial images and local ground images.}
Errors in tasks such as view localization and trajectory reconstruction indicate that models struggle to accurately map local ground observations onto the global layout in aerial images. Since the two viewpoints differ significantly in scale, orientation, and appearance, models need to establish stable correspondences between local visual cues and global spatial structure. Human evaluators, by contrast, can more effectively exploit shared spatial structure across the two views to achieve more accurate aerial-ground scene registration, while current models remain limited in this capability.

\textbf{Limited cross-view reference-frame transformation.}
Low performance on tasks such as relative orientation determination suggests that models still face clear bottlenecks in transforming reference frames across different observing subjects. In particular, models remain unstable when converting between observer-centered directions and scene-centered spatial layouts, as well as when interpreting subject orientation and directional semantics such as front, back, left, and right. In contrast, human evaluators can usually rely on reference objects and scene structure to perform more reliable aerial-ground coordinate transformation, indicating that current models still need stronger cross-view spatial modeling capabilities.

\textbf{Insufficient modeling of visibility and occlusion.}
Errors in occlusion reasoning tasks show that models have difficulty jointly considering subject position, viewing direction, and scene obstacles to determine whether a target is visible from another viewpoint. In contrast, human evaluators can more accurately infer occlusion relations across aerial and ground views and distinguish whether a target is visible or occluded. This indicates that current models remain limited in modeling visibility changes and occlusion relations under aerial-ground viewpoints.

\section{Conclusion}

We introduced AeroGround, a simulated benchmark for evaluating VLMs in aerial-ground collaborative reasoning. Built from paired observations in diverse open environments, AeroGround provides 2,250 question-answering instances across 3 capability levels and 11 tasks, covering cross-view alignment, spatial understanding, and reasoning. Experiments on representative VLMs reveal a substantial gap between current models and human performance, showing that aerial-ground reasoning remains challenging, especially in cross-view transfer, spatial alignment, and unified scene modeling. Domain fine-tuning yields modest performance gains, further validating the value of AeroGround for model diagnosis and adaptation.

AeroGround also has several limitations that suggest directions for future work. First, although we provide a small-scale real-world pilot evaluation in Appendix~\ref{app:real-world-pilot}, the benchmark itself remains simulation-based. Future work should extend the real-world evaluation to larger and more diverse scenes while preserving controllable annotation. Second, it currently focuses on static benchmark evaluation, and future extensions can move toward dynamic and interactive aerial-ground collaboration. Third, current models still lack unified spatial intelligence, motivating future methods that combine multi-view geometry, spatial memory, and language reasoning for more robust cross-view understanding.

{
\small

\bibliographystyle{plainnat}
\bibliography{ref}
}

\newpage

\appendix

\section{Dataset and Benchmark Details}
\label{app:dataset-benchmark-details}

\subsection{Overview of AeroGround}

AeroGround consists of a simulated aerial--ground dataset and a QA benchmark built from paired aerial--ground observations. The dataset contains 29,177 simulated multimodal observations, each consisting of an RGB image and simulator-derived metadata, including depth, camera pose, and instance annotations. The benchmark contains 2,250 QA instances across 3 capability levels and 11 task categories.

This appendix provides additional details on data generation, dataset composition, benchmark taxonomy, and distribution statistics. The benchmark is organized along four dimensions: task category, directional subject, view configuration, and spatial span.

\subsection{Dataset and Benchmark Generation Pipeline}

The construction pipeline starts from simulated data acquisition and produces metadata-grounded QA instances. Each stage exports intermediate files, including RGB images, depth point clouds, pose files, instance-segmentation maps, frame-level match files, and pair-level metadata, making the pipeline extensible to new scenes, camera settings, and task generators.

\paragraph{Data acquisition.}
We use Unreal Engine 4.27 with AirSim to collect aerial and ground observations. The aerial agent uses front-view and top-down cameras, while the ground agent uses front, left, and right cameras. For each sampled frame, we save RGB images, depth point clouds, instance masks, subject poses, and camera poses. We also maintain a global instance registry containing each target's identity, category, segmentation ID, and 3D pose.

\paragraph{Single-frame instance annotation.}
Each raw frame is converted into globally aligned instance annotations. We decompose the instance mask into connected components, retrieve candidate global identities, project candidate 3D positions into the image, and score candidates using mask, center, depth, size, and category cues. Hungarian matching then produces one-to-one assignments between visible components and global instance IDs.

\begin{compactalgo}{Algorithm A.1: Single-frame Global Instance Annotation}
  \Statex \textbf{Input:} $I,D,S,P,K,\mathcal{R}$
  \Statex \textbf{Output:} frame annotation $\mathcal{M}_f$
  \State $\mathcal{C} \gets \mathrm{CC}(S)$
  \For{$c_i \in \mathcal{C}$}
    \State $\phi_i \gets (m_i,b_i,\bar{\mathbf{x}}_i,a_i,\rho_i,\bar{d}_i)$
    \State $\mathcal{U}_i \gets \mathcal{R}[\mathrm{seg}(c_i)]$
    \For{$u \in \mathcal{U}_i$}
      \State $\hat{\mathbf{x}}_{iu},\hat{d}_{iu} \gets \pi_K(P^{-1}\mathbf{p}_u)$
      \State $s_{iu} \gets w_m\mathbf{1}[\hat{\mathbf{x}}_{iu}\in m_i] + w_c g(\hat{\mathbf{x}}_{iu},\bar{\mathbf{x}}_i)$
      \State $s_{iu} \gets s_{iu}+w_d g(\hat{d}_{iu},\bar{d}_i)+w_a h(a_i,\hat{d}_{iu})+w_g h(\rho_i,\mathrm{cat}_u)$
    \EndFor
  \EndFor
  \State $\mathcal{A}^{\star}\gets \arg\max_{\mathcal{A}\in\Omega}\sum_{(i,u)\in\mathcal{A}}s_{iu}$
  \State $\mathcal{M}_f \gets \{(c_i,u,s_{iu},\phi_i):(i,u)\in\mathcal{A}^{\star}\}$
\end{compactalgo}

\paragraph{Aerial--ground view pairing.}
Frame-level match files enable automatic pairing using shared global identities and camera geometry. For each candidate pair, we check shared instances, camera distance, scene-center distance, and viewing direction. Valid pairs are scored and labeled with view configuration and spatial span.

\begin{compactalgo}{Algorithm A.2: Automatic Aerial-Ground View Pairing}
  \Statex \textbf{Input:} $\mathcal{F}_a,\mathcal{F}_g,\mathcal{R},\tau_s,\tau_{xy},\tau_c,\tau_\theta$
  \Statex \textbf{Output:} pair set $\mathcal{P}$
  \State $\mathcal{P}\gets\emptyset$
  \State $\mathcal{I}(u)\gets\{g\in\mathcal{F}_g:u\in\mathcal{U}_g\}$
  \For{$a\in\mathcal{F}_a$}
    \State $\mathcal{G}_a \gets \bigcup_{u\in\mathcal{U}_a}\mathcal{I}(u)$
    \For{$g\in\mathcal{G}_a$}
      \State $\mathcal{S}_{a,g}\gets\mathcal{U}_a\cap\mathcal{U}_g$
      \State $\mathbf{c}_{a,g}\gets|\mathcal{S}_{a,g}|^{-1}\sum_{u\in\mathcal{S}_{a,g}}\mathbf{p}_u$
      \If{$|\mathcal{S}_{a,g}|<\tau_s$ or $d_{xy}(a,g)>\tau_{xy}$}
        \State \textbf{continue}
      \EndIf
      \State $\theta_{a,g}\gets\max(\angle(\mathbf{f}_a,\mathbf{c}_{a,g}-\mathbf{o}_a),\angle(\mathbf{f}_g,\mathbf{c}_{a,g}-\mathbf{o}_g))$
      \If{$\max(d(a,\mathbf{c}_{a,g}),d(g,\mathbf{c}_{a,g}))>\tau_c$ or $\theta_{a,g}>\tau_\theta$}
        \State \textbf{continue}
      \EndIf
      \State $q_{a,g}\gets \alpha|\mathcal{S}_{a,g}|+\beta\bar{s}_{a,g}+\gamma\kappa_{a,g}+\eta r_{a,g}$
      \State $q_{a,g}\gets q_{a,g}+\lambda(1-\theta_{a,g}/\tau_\theta)$
      \State $v_{a,g}\gets\mathrm{ViewLabel}(a,g)$
      \State $\ell_{a,g}\gets\mathrm{SpanLabel}(d_{xy}(a,g))$
      \State $\mathcal{P}\gets\mathcal{P}\cup\{(a,g,\mathcal{S}_{a,g},q_{a,g},v_{a,g},\ell_{a,g})\}$
    \EndFor
  \EndFor
\end{compactalgo}

\paragraph{QA generation and filtering.}
The retained pairs are used to instantiate task-specific question templates. Ground-truth answers are computed from structured metadata. We remove samples with insufficient visual evidence, unstable visibility, non-unique answers, invalid candidates, ambiguous wording, unreliable pairing, or multiple plausible answers. The final benchmark contains 2,250 high-quality QA instances.

\subsection{Simulated Dataset Composition}

The simulated data pool covers cities, towns, factories, deserts, and forests. It contains 29,177 RGB observations, including 15,335 aerial-view images and 13,842 ground-view images. Table~\ref{tab:simulated-dataset-composition} reports the raw RGB image statistics by scene type; derived folders such as paired outputs and benchmark files are excluded.

\begin{table}[!htbp]
  \centering
  \small
  \setlength{\tabcolsep}{10pt}
  \caption{Composition of the raw simulated AeroGround data pool by scene type.}
  \label{tab:simulated-dataset-composition}
  \vspace{0.3em}
  \begin{tabular}{lrrr}
    \toprule
    Scene type & Aerial & Ground & Total \\
    \midrule
    City    & 8,686  & 6,756  & 15,442 \\
    Town    & 5,664  & 4,499  & 10,163 \\
    Factory & 248    & 825    & 1,073  \\
    Desert  & 491    & 355    & 846    \\
    Forest  & 246    & 1,407  & 1,653  \\
    \midrule
    \textbf{Total} & \textbf{15,335} & \textbf{13,842} & \textbf{29,177} \\
    \bottomrule
  \end{tabular}
\end{table}

Each observation is paired with simulator metadata, including camera poses, depth point clouds, instance masks, global instance IDs, and object-level states. These annotations support cross-view matching, visibility checking, occlusion reasoning, and trajectory-based QA generation.

\subsection{AeroGround Benchmark Composition}

The AeroGround Benchmark contains 2,250 QA instances generated from paired aerial--ground observations. Each instance stores the involved views, target objects, answer candidates, ground-truth answer, task category, capability level, directional subject type, view configuration, and spatial span. The benchmark uses closed-form or format-constrained answers for automatic evaluation.

For domain fine-tuning, we use 9,475 high-quality training samples constructed from scene assets disjoint from the benchmark test set. This split reduces scene leakage and evaluates transferable aerial--ground reasoning.

AeroGround evaluates 3 capability levels and 11 task categories, following Figure~\ref{fig:benchmark-design}(c). Table~\ref{tab:task-taxonomy} lists the task categories and their core question templates.

\begin{table}[!htbp]
  \centering
  \scriptsize
  \setlength{\tabcolsep}{3pt}
  \caption{Task taxonomy and core question templates of the AeroGround Benchmark. Task names are consistent with Figure~\ref{fig:benchmark-design}(c).}
  \label{tab:task-taxonomy}
  \vspace{0.3em}
  \begin{tabular}{p{0.18\linewidth} p{0.25\linewidth} p{0.50\linewidth}}
    \toprule
    Capability level & Task category & Core question template \\
    \midrule

    \multirow[c]{2}{=}[-0.35em]{Object Alignment}
    & Single Target Matching
    & Select the candidate target in [target view] that corresponds to the marked target in [source view]. \\

    & Multi-Target Matching
    & Determine the one-to-one correspondence from $A_1,\ldots,A_k$ to the correct candidate labels. \\

    \midrule

    \multirow[c]{5}{=}[-0.75em]{Spatial Understanding}
    & Relative Distance Comparison
    & Order $T_1,\ldots,T_k$ from nearest to farthest relative to [agent position]. \\

    & Multi-Target Relation
    & Order the comparison targets from nearest to farthest relative to reference target $R$. \\

    & Region Counting
    & Count how many [category] instances lie inside the marked aerial region. \\

    & Relative Orientation Determination
    & Select the egocentric quadrant of [aerial/teammate direction] relative to the ground agent's facing direction. \\

    & View Localization
    & Select the aerial grid cell or the point that contains the ground-view capture position. \\

    \midrule

    \multirow[c]{4}{=}[-0.60em]{Spatial Reasoning}
    & Occluded Target Judgment
    & Determine whether the same target is visible, occluded, or outside the field of view in [target view]. \\

    & View Transformation Reasoning
    & Select the candidate ground view that matches the marked position and facing direction in the aerial view. \\

    & Reverse Sequence Reasoning
    & Identify the candidate ground frame that does not belong to the true trajectory sequence. \\

    & Trajectory Reconstruction Selection
    & Select the aerial trajectory that best matches the true path of the ground-view sequence. \\

    \bottomrule
  \end{tabular}
\end{table}

\begin{figure}[!htbp]
  \centering
  \includegraphics[width=0.72\linewidth]{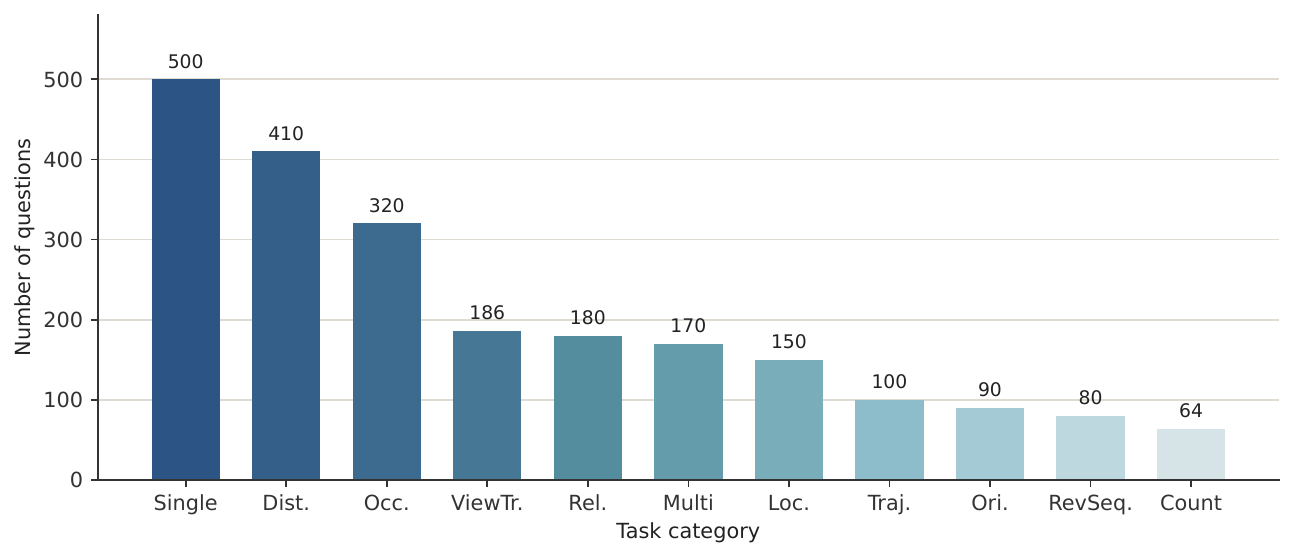}
  \caption{Task category distribution of the AeroGround Benchmark.}
  \label{fig:task-category-distribution}
\end{figure}

Figure~\ref{fig:task-category-distribution} shows the distribution of QA instances across the 11 task categories.

\subsection{View Configuration, Directional Subject, and Spatial Span Statistics}

Directional subject describes the answer anchor and cross-view information flow. \textbf{Aerial-to-Ground} tasks anchor the answer in the ground view using aerial evidence, \textbf{Ground-to-Aerial} tasks anchor the answer in the aerial view using ground evidence, and \textbf{Non-directional} tasks do not require a dominant transfer direction. Single Target Matching, Multi-Target Matching, and Occluded Target Judgment contain both Aerial-to-Ground and Ground-to-Aerial subtypes.

View configuration describes the geometry of aerial--ground observations. Aerial views include top-down and front-view settings; front-view cases are further divided into same-side and opposite-side configurations. We define a front-view pair as same-side when the angle between the aerial and ground camera viewing directions is less than $45^\circ$; otherwise, it is treated as opposite-side. Spatial span is measured by the relative 3D distance between aerial and ground subjects.

\begin{figure}[!htbp]
  \centering
  \begin{minipage}{0.48\linewidth}
    \centering
    \includegraphics[width=\linewidth]{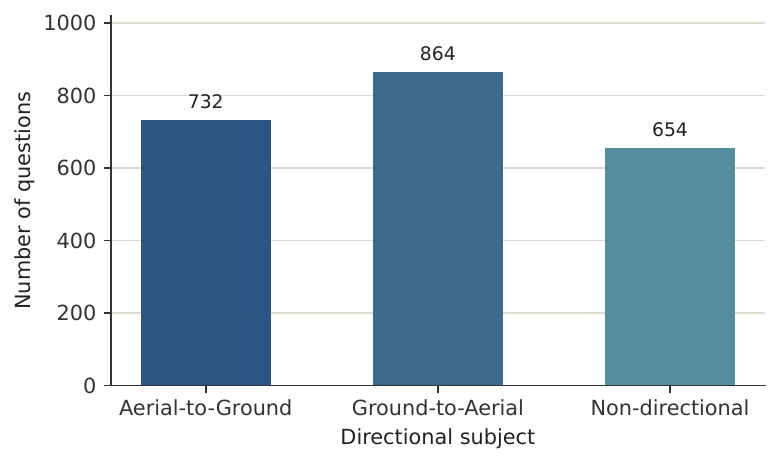}
    \vspace{-0.4em}
    \centerline{\scriptsize (a) Directional subject distribution.}
  \end{minipage}
  \hfill
  \begin{minipage}{0.48\linewidth}
    \centering
    \includegraphics[width=\linewidth]{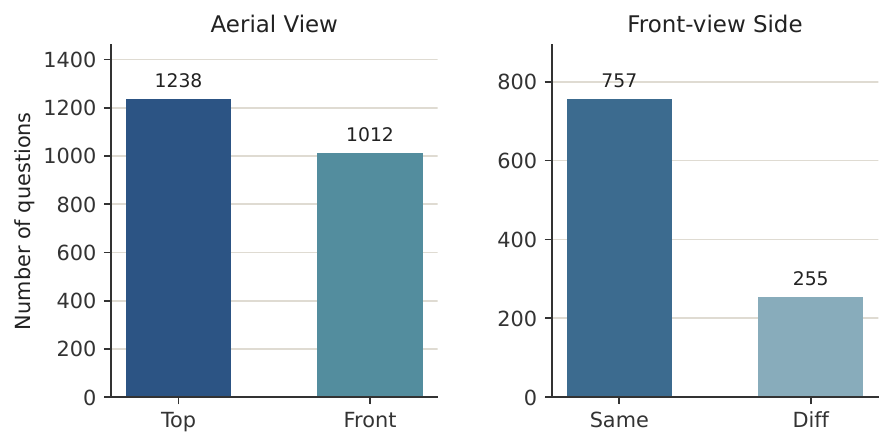}
    \vspace{-0.4em}
    \centerline{\scriptsize (b) View configuration distribution.}
  \end{minipage}

  \vspace{0.5em}

  \begin{minipage}{0.66\linewidth}
    \centering
    \includegraphics[width=\linewidth]{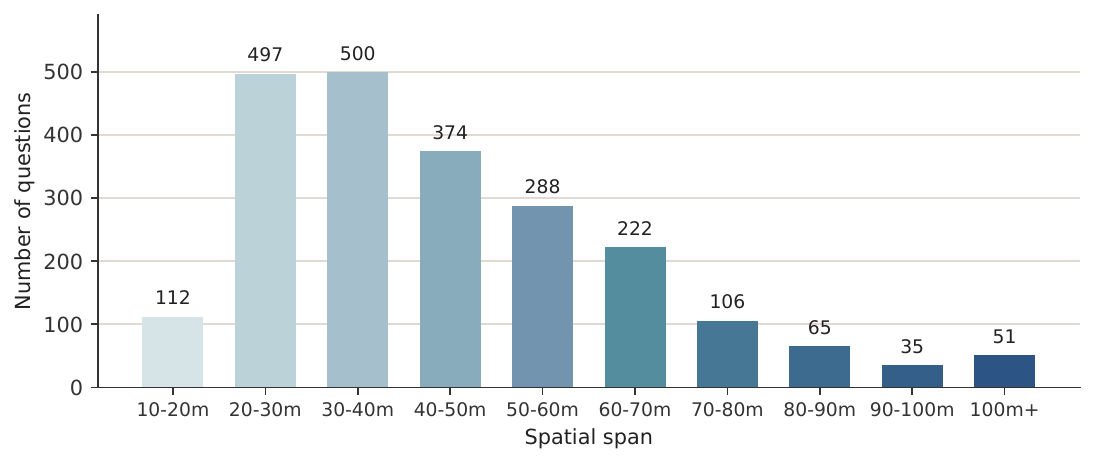}
    \vspace{-0.4em}
    \centerline{\scriptsize (c) Spatial span distribution.}
  \end{minipage}

  \vspace{0.2em}
  \caption{Statistics of aerial--ground viewpoint relations in AeroGround.}
  \label{fig:view-direction-span-statistics}
  \vspace{-0.6em}
\end{figure}

Figure~\ref{fig:view-direction-span-statistics} summarizes the distributions of directional subject types, view configurations, and spatial spans.

\section{Experiment Protocol Details}
\label{app:evaluation-protocol}

\subsection{Evaluated Models and Inference Setup}
\label{app:evaluated-models-inference}

We evaluate the models reported in the main experimental tables, including closed-source VLMs, open-source VLMs, spatial-reasoning-enhanced models, and Qwen3-VL variants used for domain fine-tuning analysis. All models use the same image inputs, question text, candidate options, and output constraints. Closed-source models are evaluated through public API endpoints,whereas most open-source models are evaluated locally using officialcheckpoints or publicly released weights. Kimi-K2.5 is an open-source model but is evaluated through the official Moonshot API. Exact API access dates and checkpoint versions will be included in the released evaluation logs.

Each benchmark instance is converted into a standardized multimodal prompt specifying the aerial and ground views, answering subject, task question, and allowed answer format. Directional tasks explicitly state which view provides evidence and which view anchors the answer. The prompt preserves the asymmetric aerial--ground observation setting rather than treating all images as unordered visual context.

Most tasks are multiple-choice questions and require the model to output one option letter. Multi-Target Matching is the only exception: it requires a complete one-to-one mapping between source targets and candidate labels. Verbose explanations are prohibited, and the model is instructed to output only the final answer.

Unless otherwise required by a model interface, we use deterministic decoding settings. For API models whose decoding parameters are not fully exposed, we use the most deterministic available setting. The same prompt template and image ordering are used for all models.

For API-evaluated models, including the closed-source models and
Kimi-K2.5, the underlying computation is handled by the corresponding
providers. The remaining open-source models are evaluated locally on
GPU machines equipped with two NVIDIA RTX 4090 GPUs and one NVIDIA H100
GPU. Each model is evaluated once on the same 2,250 benchmark questions.
The computational cost is dominated by model inference, and no extensive
hyperparameter search is performed for benchmark evaluation.

\subsection{Answer Parsing and Scoring}
\label{app:answer-parsing-scoring}

AeroGround uses closed-form evaluation. For multiple-choice tasks, the parser extracts a single valid option letter from the model response and compares it with the ground-truth option. Responses with no valid option, conflicting options, or ambiguous extra text are marked as incorrect.

Multi-Target Matching is scored by strict template matching. The parser normalizes separators, spaces, and letter cases, then extracts the predicted one-to-one correspondence. A prediction is correct only when the full mapping exactly matches the metadata-derived ground truth; no partial credit is assigned.

Table~\ref{tab:answer-parsing-rules} summarizes the scoring rules.

\vspace{0.4em}
\begin{table}[!htbp]
  \centering
  \scriptsize
  \setlength{\tabcolsep}{4pt}
  \caption{Answer parsing and scoring rules used in AeroGround.}
  \label{tab:answer-parsing-rules}
  \vspace{0.3em}
  \begin{tabular}{p{0.28\linewidth} p{0.34\linewidth} p{0.30\linewidth}}
    \toprule
    Output format & Parsing rule & Correctness criterion \\
    \midrule
    Multiple-choice option
    & Extract one valid option letter from the allowed option set.
    & Correct only if the option equals the ground-truth option. \\

    Multi-target matching template
    & Normalize the response and extract the complete one-to-one correspondence.
    & Correct only if the full mapping exactly matches the ground-truth mapping. \\

    Invalid or ambiguous response
    & No valid answer, multiple conflicting answers, incomplete mapping, or unconstrained explanation.
    & Always scored as incorrect. \\
    \bottomrule
  \end{tabular}
\end{table}
\vspace{0.2em}

For a question set $\mathcal{Q}$, the primary evaluation metric is exact-match accuracy:
\begin{equation}
  \mathrm{Acc}(\mathcal{Q}) =
  \frac{1}{|\mathcal{Q}|}
  \sum_{i \in \mathcal{Q}}
  \mathbb{I}\left[\mathrm{Parse}(r_i)=y_i\right],
\end{equation}
where $r_i$ is the model response and $y_i$ is the ground-truth answer. All main tables and primary analyses report raw accuracy unless otherwise specified.

We also compute random-choice baselines to interpret task difficulty. For multiple-choice tasks, the baseline is computed from the number of valid options. For Multi-Target Matching, it is computed from the number of valid one-to-one mappings.

For analyses where groups may differ in answer-space size or task mixture, we report chance-adjusted accuracy (CAA):
\begin{equation}
  \mathrm{CAA}(\mathcal{Q}) =
  \frac{\mathrm{Acc}(\mathcal{Q}) - R(\mathcal{Q})}
       {1 - R(\mathcal{Q})},
\end{equation}
where $R(\mathcal{Q})$ is the random-choice baseline for the corresponding question group. CAA is used only as a supplementary metric in multidimensional analyses.

\subsection{Supervised Fine-tuning Experiment}
\label{app:sft-details}

We conduct supervised fine-tuning on the scene-disjoint AeroGround training split described in Appendix~\ref{app:dataset-benchmark-details}. The split contains 9,475 QA samples generated from scene assets that do not overlap with the benchmark test set.

All Qwen3-VL fine-tuning experiments are conducted on a single NVIDIA RTX 4090 GPU. We use parameter-efficient LoRA fine-tuning with the SWIFT framework. LoRA modules are injected into eligible linear layers, while the visual encoder and visual-language alignment layers are frozen to preserve pretrained visual representations and reduce memory cost.

Table~\ref{tab:sft-hyperparameters} summarizes the main hyperparameters. We use AdamW, a peak learning rate of $1\times10^{-5}$, cosine scheduling, and a warmup ratio of 0.05. The physical batch size is 1, with 16 gradient accumulation steps. The maximum image size is 262,144 pixels, and the maximum sequence length is 4,096 tokens. The 2B model is trained for 2 epochs, and the 8B model for 4 epochs.

\begin{table}[!htbp]
  \centering
  \small
  \setlength{\tabcolsep}{6pt}
  \caption{Hyperparameters for supervised fine-tuning of Qwen3-VL models on AeroGround.}
  \label{tab:sft-hyperparameters}
  \vspace{0.3em}
  \begin{tabular}{ll}
    \toprule
    Item & Setting \\
    \midrule
    Fine-tuning method & LoRA with SWIFT \\
    Trainable modules & LoRA adapters on eligible linear layers \\
    Frozen modules & Visual encoder and visual-language alignment layers \\
    LoRA rank & 32 \\
    LoRA scaling factor & 64 \\
    LoRA dropout & 0.05 \\
    Optimizer & AdamW \\
    Peak learning rate & $1\times10^{-5}$ \\
    Learning-rate scheduler & Cosine annealing \\
    Warmup ratio & 0.05 \\
    Weight decay & 0.01 \\
    Physical batch size & 1 \\
    Gradient accumulation steps & 16 \\
    Effective batch size & 16 \\
    Gradient checkpointing & Enabled \\
    Maximum image pixels & 262,144 \\
    Maximum sequence length & 4,096 tokens \\
    Training epochs & 2 epochs for 2B; 4 epochs for 8B \\
    Hardware & Single NVIDIA RTX 4090 GPU \\
    \bottomrule
  \end{tabular}
\end{table}

During training, each sample follows the same multi-image input convention as benchmark evaluation. The model is supervised to output only the required answer format, such as an option letter, an ordered list, or a constrained categorical answer.

\subsection{Human Evaluation Details}
\label{app:human-evaluation}

We conduct human evaluation on the entire AeroGround Benchmark.
Four research colleagues independently complete all 2,250 questions
across the 11 task categories, providing a full-benchmark human-level
reference. All evaluators were informed of the purpose, procedure,
and intended research use of the evaluation and provided informed
consent before participation. Participation was voluntary, and the
evaluators received no financial compensation, as they contributed
to the evaluation as research colleagues.

Evaluators use a dedicated interface that presents the same visual
inputs, task questions, candidate options, and answer constraints as
model evaluation. They receive no feedback and cannot access the
ground-truth answers during evaluation. Multiple-choice tasks require
selecting one option, while Multi-Target Matching requires completing
the one-to-one correspondence template.

Human responses are scored with the same parser and exact-match protocol used for models. We report the mean exact-match accuracy over the four evaluators:
\begin{equation}
  \mathrm{Acc}_{\mathrm{human}} =
  \frac{1}{4}
  \sum_{j=1}^{4}
  \frac{1}{|\mathcal{Q}|}
  \sum_{i \in \mathcal{Q}}
  \mathbb{I}\left[\mathrm{Parse}(r_{i}^{(j)})=y_i\right],
\end{equation}
where $r_i^{(j)}$ is evaluator $j$'s response to question $i$, and $y_i$ is the ground-truth answer. Task-level and dimension-level human results follow the same grouping protocol as model evaluation.

\section{Additional Experimental Results and Analyses}
\label{app:additional-results}

\subsection{Real-World Pilot Evaluation}
\label{app:real-world-pilot}

\paragraph{Dataset and pilot construction.}
To examine whether the spatial-capability structure measured by
AeroGround transfers beyond simulation, we construct a 300-question
real-world pilot set using observations from the Cross-View Urban
Traffic Dataset (CVUTD)~\cite{bhardwaj2026crossviewurbantrafficdataset}.
CVUTD provides synchronized egocentric and aerial drone observations
captured at real urban intersections. We use it only as the source of
real aerial--ground observations and construct AeroGround-style
question-answering instances on top of these observations; the pilot
set is therefore distinct from the original tasks defined by CVUTD.

The pilot covers seven task types shared with AeroGround:
Single Target Matching, Multi-Target Matching, View Localization,
Relative Orientation Determination, Occluded Target Judgment,
View Transformation Reasoning, and Trajectory Reconstruction Selection.
The corresponding numbers of questions are 66, 50, 30, 24, 50, 50,
and 30, respectively. The task definitions, answer formats, and
exact-match evaluation protocol are aligned with AeroGround.
All questions were manually reviewed to ensure sufficient visual
evidence, unambiguous wording, and a unique answer.

\paragraph{Distribution-aligned comparison.}
We evaluate nine representative VLMs on the real-world pilot.
Because its task distribution differs from that of the full AeroGround
test set, directly comparing the two overall accuracies would introduce
a task-mixture confound. We therefore select the same seven task types
from AeroGround and recompute a distribution-aligned simulated score:

\begin{equation}
  \mathrm{Acc}_{\mathrm{sim}}^{\mathrm{aligned}}(m)
  =
  \sum_{t \in \mathcal{T}}
  \frac{N_{t}^{\mathrm{real}}}{N^{\mathrm{real}}}
  \mathrm{Acc}_{m,t}^{\mathrm{sim}},
\end{equation}

where $\mathcal{T}$ denotes the seven shared task types,
$N_{t}^{\mathrm{real}}$ is the number of real-world questions for task
$t$, $N^{\mathrm{real}}=300$, and
$\mathrm{Acc}_{m,t}^{\mathrm{sim}}$ is the accuracy of model $m$ on
the corresponding AeroGround task.

\begin{table*}[!t]
  \caption{
  Results on the 300-question real-world pilot set.
  Aligned Sim. denotes the AeroGround score recomputed over the same
  seven tasks using the real-world task distribution.
  The remaining task columns report real-world accuracies (\%).
  Bold and underline indicate the best and second-best overall scores,
  respectively.
  }
  \label{tab:real-world-pilot}
  \centering
  \scriptsize
  \setlength{\tabcolsep}{3.2pt}
  \resizebox{\textwidth}{!}{%
  \begin{tabular}{lccccccccc}
    \toprule
    Model & Aligned Sim. & Real Avg. & Single & Multi & Loc. & Ori. & Occ. & ViewTr. & Traj. \\
    \midrule
    No. questions & -- & 300 & 66 & 50 & 30 & 24 & 50 & 50 & 30 \\
    \midrule
    Doubao-seed-2.0-pro
    & \textbf{46.8} & \textbf{45.7}
    & 59.1 & 30.0 & 40.0 & 41.7 & 56.0 & 50.0 & 26.7 \\

    Gemini-3.1-Pro
    & \underline{39.8} & \underline{44.0}
    & 56.1 & 36.0 & 36.7 & 54.2 & 60.0 & 32.0 & 23.3 \\

    Qwen3-VL-32B-Instruct
    & 29.7 & 34.3
    & 50.0 & 12.0 & 56.7 & 25.0 & 38.0 & 34.0 & 16.7 \\

    Qwen3-VL-8B-Instruct
    & 27.7 & 33.7
    & 50.0 & 10.0 & 50.0 & 0.0 & 40.0 & 36.0 & 33.3 \\

    Spatial-SSRL-Qwen3VL-4B
    & 27.6 & 31.7
    & 50.0 & 2.0 & 26.7 & 25.0 & 44.0 & 30.0 & 33.3 \\

    InternVL3.5-38B
    & 31.4 & 30.3
    & 42.4 & 4.0 & 43.3 & 37.5 & 30.0 & 32.0 & 26.7 \\

    GLM-4.1V-9B-Base
    & 28.8 & 29.7
    & 51.5 & 2.0 & 30.0 & 16.7 & 46.0 & 18.0 & 30.0 \\

    LLaVA-OneVision-7B
    & 26.4 & 26.0
    & 27.3 & 0.0 & 20.0 & 0.0 & 46.0 & 48.0 & 23.3 \\

    SpaceOm-4B
    & 24.6 & 23.3
    & 21.2 & 0.0 & 16.7 & 8.3 & 34.0 & 52.0 & 20.0 \\
    \bottomrule
  \end{tabular}}
\end{table*}

\paragraph{Cross-domain agreement.}
The aligned simulated and real-world model scores exhibit strong and
statistically significant agreement. Across the nine evaluated models,
the Spearman rank correlation is $\rho=0.817$ ($p=0.0072$), and the
Pearson correlation is $r=0.922$ ($p=0.0004$). The mean absolute
difference between aligned simulated and real-world accuracies is
2.62 percentage points.

Doubao-seed-2.0-pro and Gemini-3.1-Pro remain the two strongest models
in both settings, whereas LLaVA-OneVision-7B and SpaceOm-4B remain the
two weakest. This agreement suggests that the broad model-level
capability structure revealed by AeroGround is not solely an artifact
of particular simulated assets or rendering styles.

At the same time, several middle-ranked models change positions, and
the real-world task-level results remain strongly model dependent.
These variations indicate domain shifts caused by realistic appearance,
occlusion, viewpoint variation, and scene complexity. Therefore, the
pilot provides preliminary evidence of cross-domain validity rather
than demonstrating that the simulation-to-real gap has been eliminated.
Its limited scale and focus on urban traffic scenes also motivate
larger and more diverse real-world evaluations.

\FloatBarrier

\subsection{Human--Model Difficulty Association}

\begin{figure}[!t]
  \centering
  \includegraphics[width=0.78\linewidth]{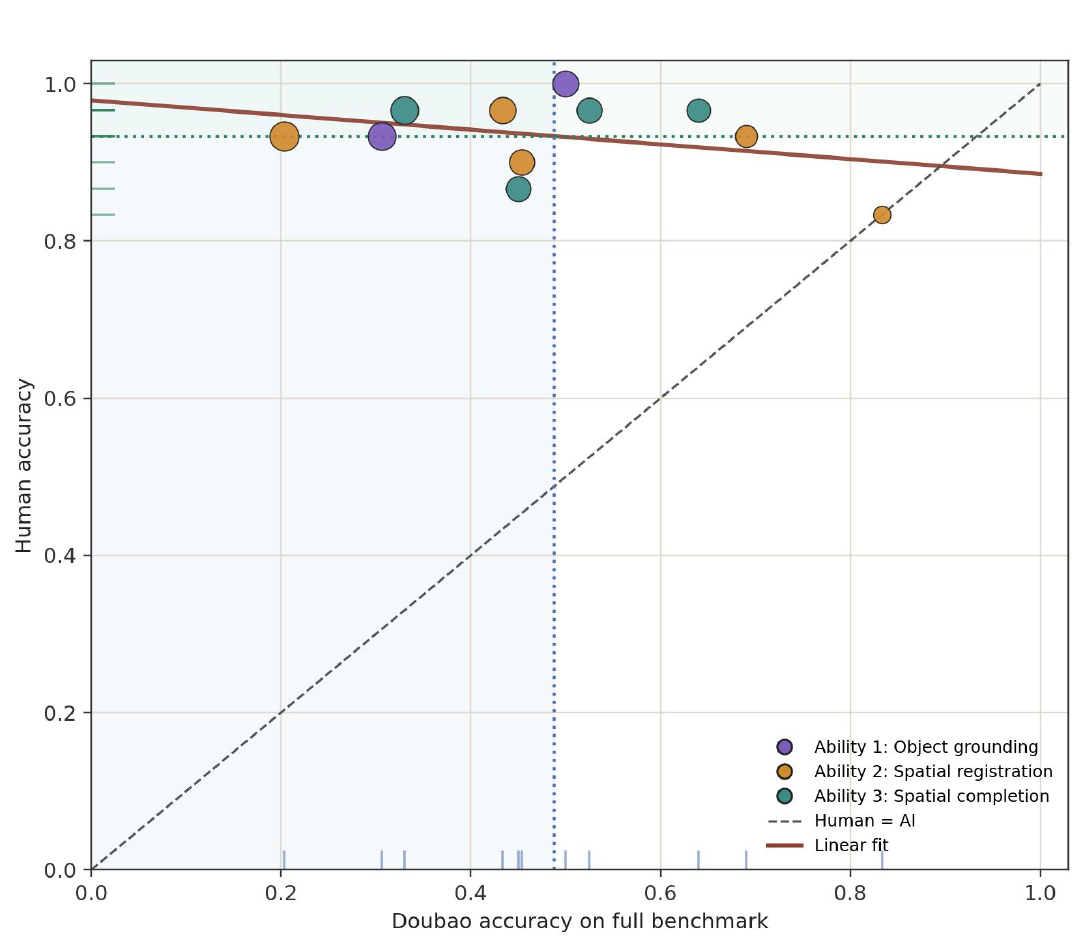}
  \caption{
  Human--model task difficulty association. Each point denotes one task, comparing human accuracy with the accuracy of Doubao-seed-2.0-pro. The plot shows that model difficulty does not fully align with human difficulty, indicating a clear gap between human spatial understanding and current model capability.
  }
  \label{fig:human-doubao-scatter}
\end{figure}

\paragraph{Analysis.}
Human performance and model performance exhibit a clear gap across tasks. More importantly, task difficulty for humans and models is not highly consistent. Some tasks that are relatively manageable for humans remain challenging for the model, suggesting that current models still lack robust aerial-ground spatial perception and cross-view reasoning abilities. This result indicates that the benchmark does not merely measure visual recognition difficulty, but also exposes model-specific limitations in open-scene spatial understanding.

\FloatBarrier

\subsection{Effect of Aerial--Ground Spatial Span}

\begin{figure}[!t]
  \centering
  \includegraphics[width=0.86\linewidth]{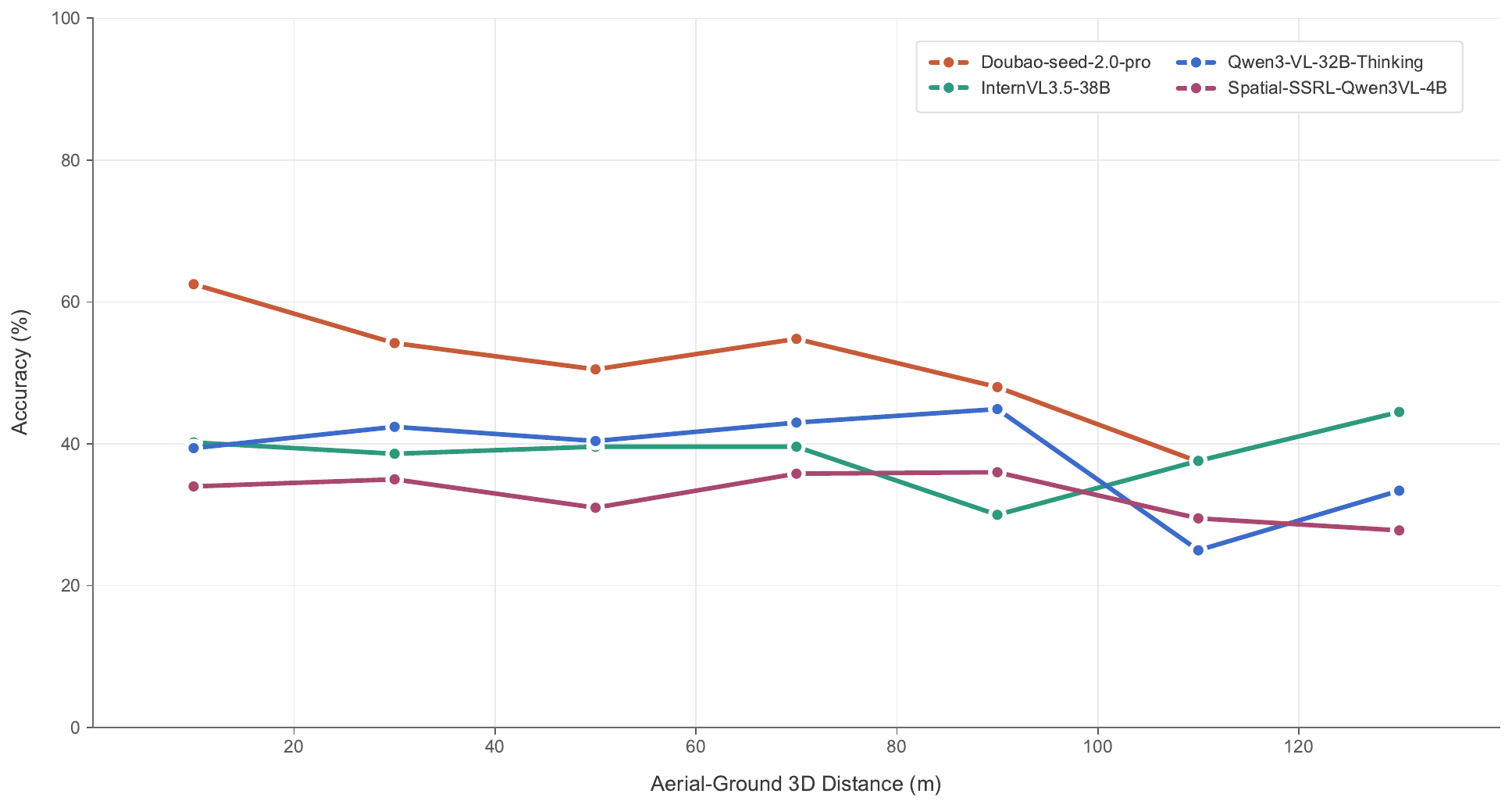}
  \caption{
  Effect of aerial--ground spatial span on model accuracy. Spatial span is measured by the 3D distance between the aerial viewpoint and the ground viewpoint. Samples are grouped into coarse 20-meter distance bins.
  }
  \label{fig:scene-distance}
\end{figure}

\paragraph{Analysis.}
Overall, a larger aerial--ground spatial span increases the difficulty of the benchmark. As the distance between the aerial and ground viewpoints grows, the ground-view region corresponds to a larger and more ambiguous candidate area in the aerial view. Meanwhile, the visual overlap between the two views becomes weaker, making cross-view spatial alignment more uncertain.

The effect of spatial span also reveals differences between local spatial reasoning and aerial--ground spatial reasoning. Tasks based on local object relations, such as multi-object relations and region counting, are relatively stable across distance changes. In contrast, tasks that require explicit aerial--ground alignment, such as view localization, are more sensitive to spatial span. This suggests that current models are more vulnerable when they need to establish correspondences between local ground observations and global aerial layouts.

\FloatBarrier

\subsection{Effect of Target Pixel Area}

\begin{figure*}[!t]
  \centering
  \begin{minipage}[t]{0.48\textwidth}
    \centering
    \includegraphics[width=\linewidth]{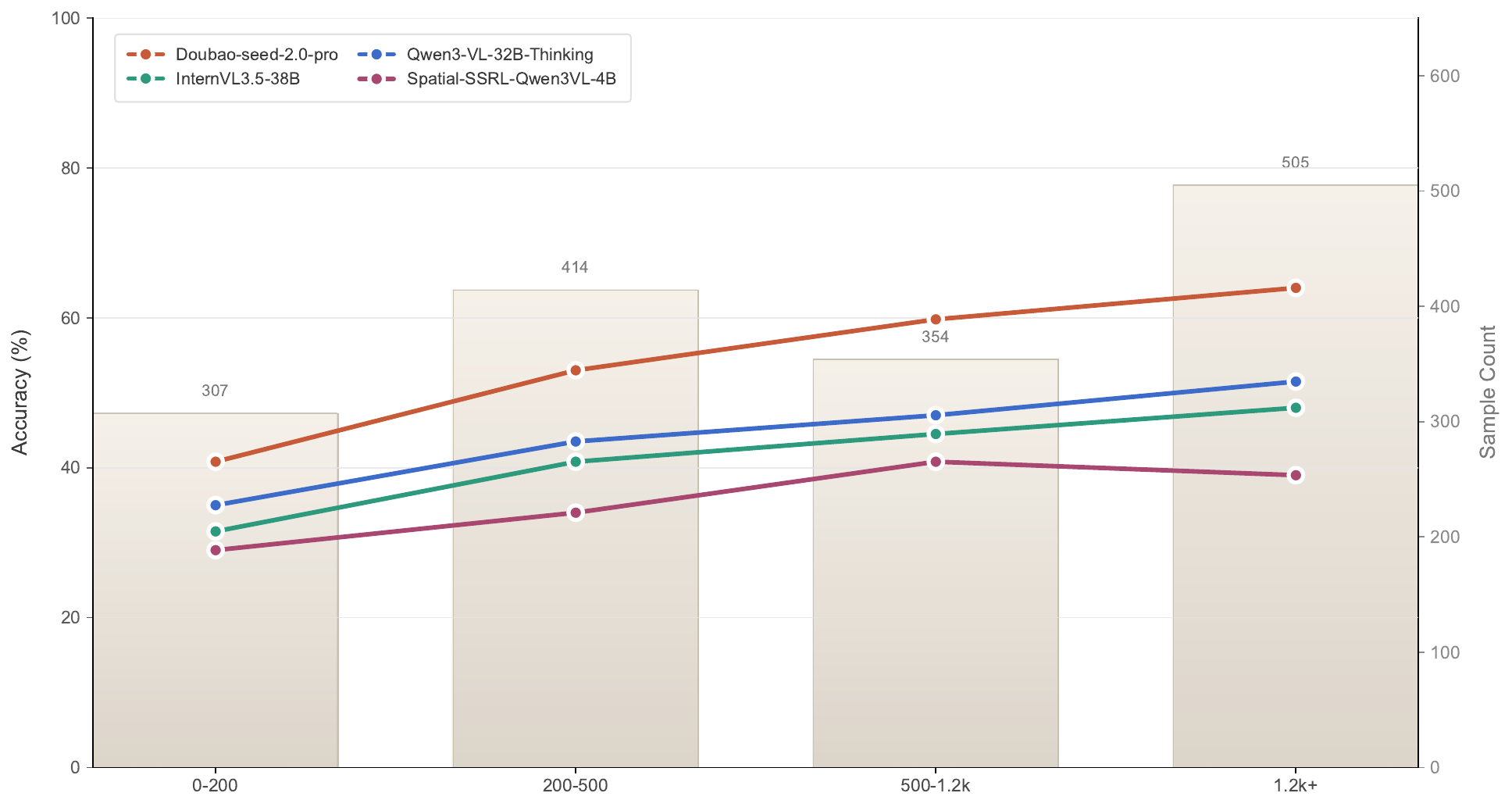}
    \vspace{0.3em}
    {\small \textbf{(a)} Overall trend.}
  \end{minipage}
  \hfill
  \begin{minipage}[t]{0.48\textwidth}
    \centering
    \includegraphics[width=\linewidth]{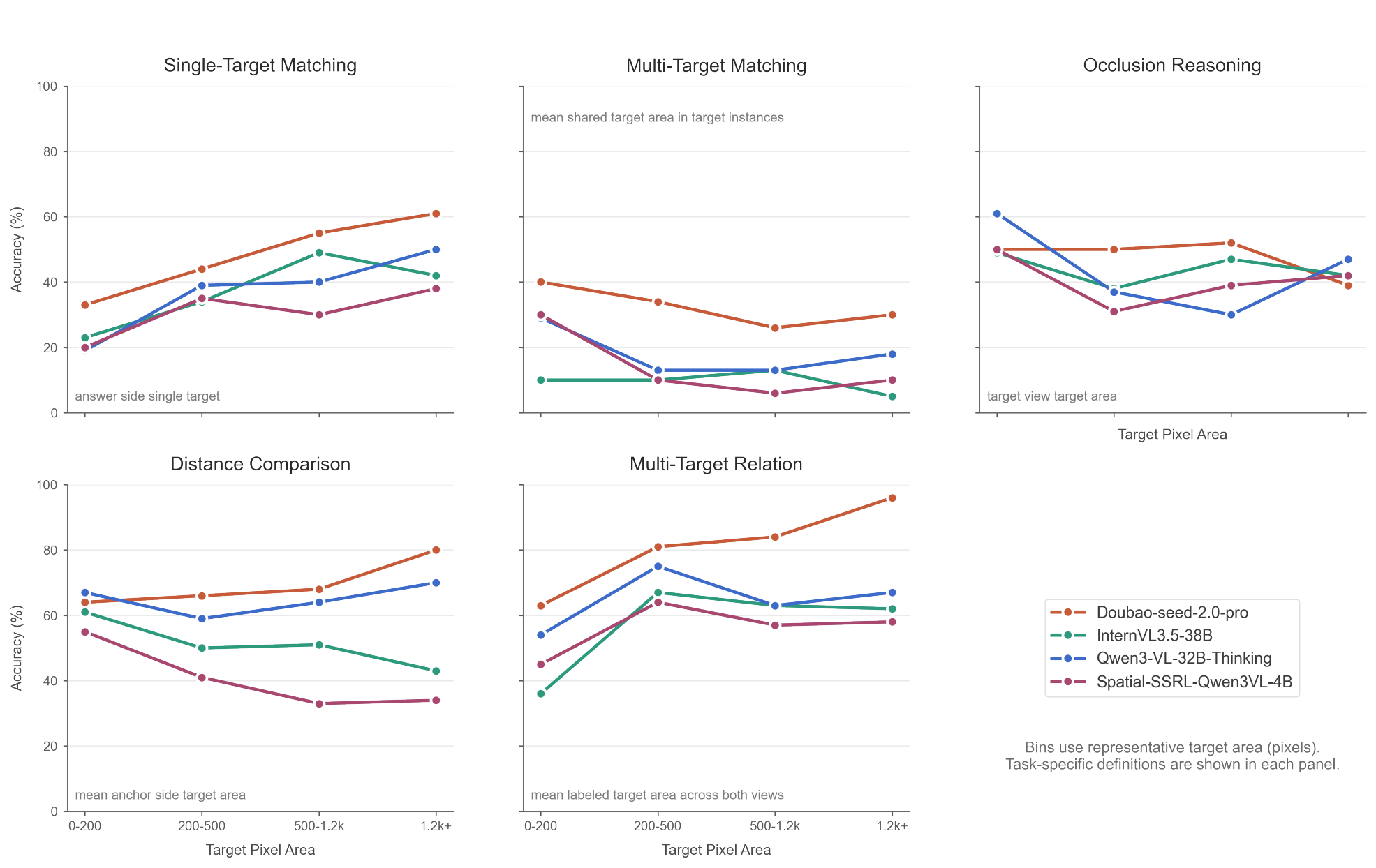}
    \vspace{0.3em}
    {\small \textbf{(b)} Task-level trend.}
  \end{minipage}

  \caption{
  Effect of target pixel area on model accuracy. We analyze five target-size-related tasks: single-object matching, multi-object matching, relative distance comparison, multi-object relation reasoning, and occluded object judgment. Left: the overall accuracy trend across target pixel area intervals. Right: task-level accuracy trends across the same intervals.
  }
  \label{fig:target-pixel-area}
\end{figure*}

\paragraph{Analysis.}
Target pixel area forms an important perceptual threshold for aerial--ground cross-view reasoning. As the target pixel area increases, model accuracy generally improves, indicating that small targets provide insufficient visual evidence and constitute a major source of failure. However, larger targets do not fully eliminate the difficulty. Even when targets occupy more pixels, several tasks remain far from saturated, suggesting that model errors also come from cross-view spatial alignment, target matching, and reasoning bottlenecks rather than visual perception alone.

\FloatBarrier

\subsection{Task Association Analysis}

\begin{figure*}[!t]
  \centering
  \includegraphics[
    width=0.72\textwidth,
    height=0.58\textheight,
    keepaspectratio,
    trim=5 5 5 17,
    clip
  ]{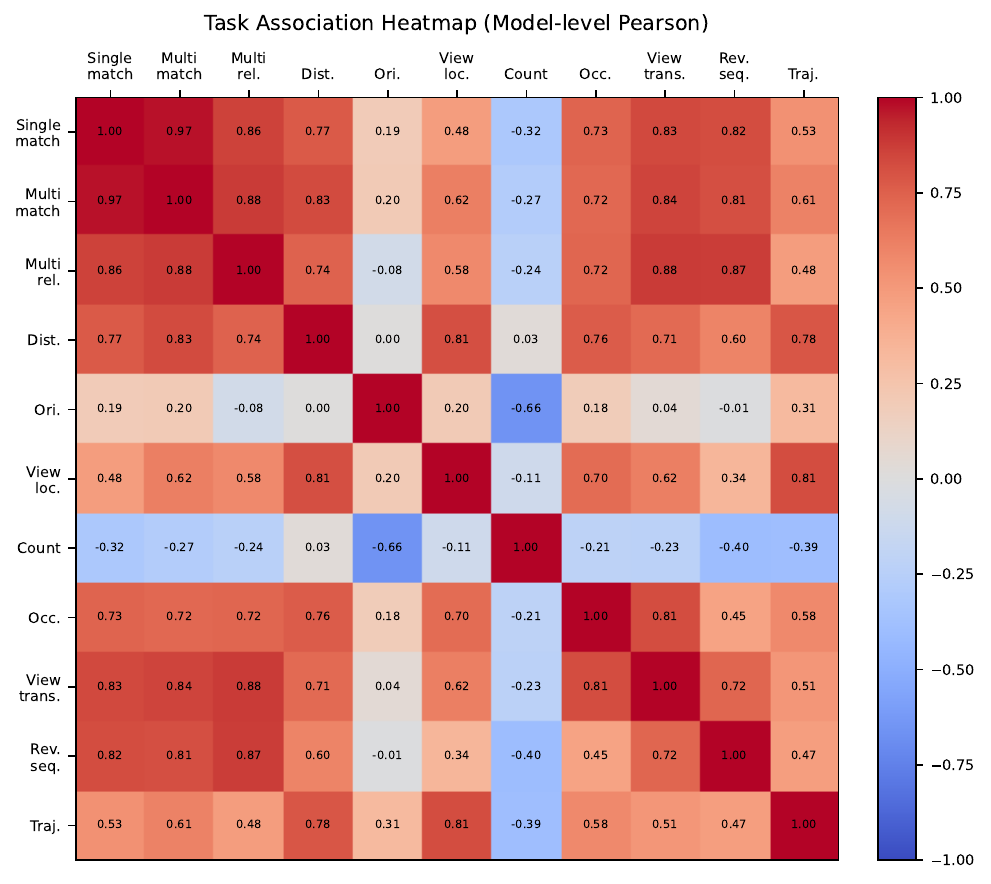}
  \caption{
  Task association analysis. The heatmap reports model-level Pearson correlations between task accuracies, revealing which task types tend to share similar model strengths and weaknesses.
  }
  \label{fig:task-association}
\end{figure*}

\paragraph{Analysis.}
Most task pairs show positive correlations. Across all 55 task pairs, the average Pearson correlation coefficient is 0.43, suggesting that the benchmark tasks are not isolated measurements but jointly depend on more fundamental aerial--ground perception, spatial alignment, and reasoning abilities.

The results further show a progressive coupling among capability levels. Object-level grounding is strongly related to spatial registration and spatial completion tasks. For example, single-object matching is highly correlated with multi-object relation reasoning and view transformation reasoning, reaching 0.86 and 0.83, respectively. View localization is also strongly correlated with trajectory selection, with a correlation of 0.81. These results indicate a hierarchical dependency from object grounding to cross-view spatial registration and then to higher-level spatial inference.

Several outlier tasks reveal more specialized spatial sub-capabilities. Relative orientation determination is less correlated with the main task cluster, suggesting that aerial--ground coordinate transformation remains a distinct challenge. Region counting also behaves as an outlier because it relies more heavily on region boundary understanding, small-object recognition, and local visual discrimination.

\FloatBarrier

\subsection{Chance-Adjusted Benchmark Results}
\label{app:caa-results}

\begin{table*}[!t]
  \caption{
  Chance-adjusted benchmark results after removing the effect of the random-choice baseline. Values are reported as CAA scores in percentage form, following the definition introduced in the evaluation protocol. A score of 0 corresponds to random-choice performance, while negative values indicate performance below the random baseline. Bold and underline indicate the best and second-best task scores within each model group, respectively.
  }
  \label{tab:caa-results}
  \centering
  \scriptsize
  \setlength{\tabcolsep}{2.4pt}
  \resizebox{\textwidth}{!}{%
  \begin{tabular}{lcccccccccccc}
    \toprule
    & & \multicolumn{2}{c}{Obj. Align.} & \multicolumn{5}{c}{Spatial Underst.} & \multicolumn{4}{c}{Spatial Reason.} \\
    \cmidrule(lr){3-4} \cmidrule(lr){5-9} \cmidrule(lr){10-13}
    Model & Avg. & Single & Multi & Rel. & Dist. & Count & Ori. & Loc. & Occ. & ViewTr. & RevSeq. & Traj. \\
    \midrule
    \multicolumn{13}{l}{\textit{Reference}} \\
    Human Level & 91.1 & 100.0 & 93.3 & 74.3 & 91.3 & 92.1 & 95.7 & 87.1 & 79.7 & 95.1 & 95.7 & 95.7 \\
    \midrule
    \multicolumn{13}{l}{\textit{Closed-source models}} \\
    GPT-5.4 & 20.6 & \underline{18.1} & 14.7 & 52.2 & 33.3 & \textbf{27.8} & \underline{8.7} & 19.8 & \underline{11.9} & 1.6 & \underline{24.1} & \underline{3.9} \\
    Gemini-3.1-Pro & \underline{30.8} & 17.5 & \underline{25.9} & \underline{67.5} & \textbf{62.5} & \underline{9.2} & \underline{8.7} & \textbf{33.6} & 3.4 & \textbf{56.8} & 16.1 & -2.6 \\
    Doubao-seed-2.0-pro & \textbf{39.5} & \textbf{34.6} & \textbf{30.6} & \textbf{74.3} & \underline{59.6} & 5.6 & \textbf{26.1} & \underline{29.2} & \textbf{16.2} & \underline{46.4} & \textbf{71.0} & \textbf{28.6} \\
    \midrule
    \multicolumn{13}{l}{\textit{Open-source models}} \\
    Kimi-K2.5 & \underline{19.5} & \textbf{21.2} & \textbf{17.6} & 28.2 & \underline{46.0} & 25.9 & -15.9 & 14.6 & 5.6 & -4.0 & -22.6 & \textbf{39.0} \\
    Qwen3-VL-8B-Thinking & 5.6 & 5.8 & 10.0 & 27.4 & 17.7 & \underline{27.8} & -20.2 & -7.8 & -9.1 & 4.8 & -14.6 & -2.6 \\
    InternVL3.5-8B & 8.5 & 15.2 & 8.8 & 15.4 & 18.4 & 24.1 & \underline{5.9} & -5.2 & -0.5 & -11.2 & -11.2 & 7.8 \\
    GLM-4.1V-9B-Base & 13.1 & 11.5 & 5.3 & 10.3 & 32.3 & 25.9 & \textbf{14.5} & 12.9 & \underline{9.0} & -6.4 & -6.5 & \underline{13.0} \\
    LLaVA-OneVision-7B & 7.8 & 5.0 & 1.2 & 23.1 & 14.2 & 22.3 & 4.4 & 10.3 & 3.4 & 8.8 & -9.7 & 1.3 \\
    Qwen3-VL-32B-Instruct & 17.0 & 13.1 & 8.8 & 25.7 & 35.8 & 22.3 & -5.7 & \textbf{19.8} & 6.7 & 8.0 & \textbf{35.5} & 5.2 \\
    Qwen3-VL-32B-Thinking & \textbf{22.5} & 17.8 & \underline{14.1} & \textbf{52.2} & \textbf{51.4} & \textbf{35.2} & -15.9 & \underline{15.5} & 4.7 & \underline{11.2} & \underline{25.8} & 3.9 \\
    InternVL3.5-38B & 18.8 & \underline{19.6} & 10.0 & \underline{42.8} & 34.9 & 24.1 & -8.6 & 2.6 & \textbf{12.8} & \textbf{23.2} & 6.5 & -1.3 \\
    \midrule
    \multicolumn{13}{l}{\textit{Spatial reasoning models}} \\
    Spatial-SSRL-Qwen3VL-4B & \textbf{12.2} & \textbf{15.4} & \textbf{7.6} & \textbf{39.4} & \underline{18.4} & 1.9 & \underline{7.3} & \underline{0.0} & \textbf{3.4} & \textbf{0.7} & \textbf{27.5} & \underline{0.0} \\
    SpaceOm-4B & \underline{7.3} & 5.8 & \underline{0.0} & \underline{-2.6} & \textbf{29.2} & \underline{14.8} & \textbf{14.5} & \textbf{11.1} & \textbf{3.4} & -16.8 & -16.1 & \textbf{7.8} \\
    SpaceThinker-Qwen2.5VL-3B & 2.7 & \underline{7.6} & \underline{0.0} & -13.7 & 13.9 & \textbf{22.3} & \underline{7.3} & -6.1 & \underline{-2.4} & \underline{-8.8} & \underline{-6.5} & -2.6 \\
    \midrule
    \multicolumn{13}{l}{\textit{Domain Fine-tuned \& Baselines}} \\
    Qwen3-VL-2B-Instruct & 6.4 & 9.4 & 2.9 & \textbf{24.8} & 14.8 & 7.5 & 0.0 & 2.6 & -0.5 & -7.1 & 1.7 & -7.8 \\
    Qwen3-VL-2B-Instruct-LoRA & 8.4 & \textbf{13.9} & 0.0 & 12.8 & \underline{22.1} & 0.0 & \underline{3.0} & \underline{8.5} & 0.9 & \underline{1.6} & -8.1 & -7.8 \\
    Qwen3-VL-8B-Instruct & \underline{11.5} & \underline{11.0} & \textbf{10.0} & \underline{22.2} & 19.4 & \underline{22.3} & \textbf{5.9} & 1.7 & \textbf{12.8} & 0.0 & \textbf{16.1} & \underline{-6.5} \\
    Qwen3-VL-8B-Instruct-LoRA & \textbf{14.5} & \textbf{13.9} & \underline{7.7} & 18.0 & \textbf{24.5} & \textbf{35.2} & 1.4 & \textbf{22.4} & \underline{5.6} & \textbf{10.4} & \underline{14.6} & \textbf{2.6} \\
    \bottomrule
  \end{tabular}}
\end{table*}

\paragraph{Analysis.}
Table~\ref{tab:caa-results} shows that the overall ranking remains largely consistent after adjusting for random-choice baselines. Human performance is still far above all model results, while closed-source models achieve the strongest overall CAA scores. Among them, Doubao-seed-2.0-pro performs best on average, and Qwen3-VL-32B-Thinking obtains the highest average score among open-source models.

The adjusted scores also highlight that many models only exceed chance level by a limited margin. Several spatial reasoning tasks, especially orientation determination, view transformation, reverse sequence reasoning, and trajectory selection, still show low or even negative CAA scores for some models. This suggests that raw accuracy can partially hide chance-level effects, and that robust aerial--ground spatial reasoning remains a major challenge.

Fine-tuned models show moderate improvements in some cases, such as Qwen3-VL-8B-Instruct-LoRA over its base counterpart, but the gains are uneven across tasks. This indicates that domain fine-tuning improves benchmark adaptation, yet does not fully resolve the deeper challenges of cross-view spatial alignment and inference.

\FloatBarrier

\section{Assets, Release, and Responsible Use}
\label{app:assets-release-responsible-use}

\subsection{Assets, Licenses, and Documentation}
\label{app:assets-licenses-documentation}

AeroGround uses existing third-party simulation assets, software,
and datasets where applicable. We credit the original creators and
follow their corresponding licenses and terms of use.
Table~\ref{tab:existing-assets-licenses} summarizes the major
third-party resources used in this work.benchmark construction.

\begin{table*}[!t]
  \centering
  \small
    \caption{Summary of major third-party assets, software, and datasets used in AeroGround.}
  \label{tab:existing-assets-licenses}
  \resizebox{\textwidth}{!}{%
  \begin{tabular}{llll}
    \toprule
    Asset & Creator / Owner & Usage & License / Terms \\
    \midrule
    Unreal Engine & Epic Games & Simulation and rendering engine & Unreal Engine EULA \\
    AirSim & Microsoft & Vehicle and camera simulation plugin/API & MIT License \\
    \midrule
    Brushify - Urban Buildings Pack & Brushify Ltd. & Urban building and environment assets & Epic Content License Agreement / UE Marketplace Standard License \\
    Downtown West Modular Pack & PurePolygons & Modular urban scene assets & Epic Content License Agreement / UE Marketplace Standard License \\
    Factory Pack Vol.1 & MeikWModels & Industrial factory scene assets & CC BY 4.0 \\
    Metaverse Villa City Modular & ErikGames & Modular villa/city scene assets & Epic Content License Agreement / UE Marketplace Standard License \\
    Modern City Bundle & Pedja Racan & Urban building and city assets & Epic Content License Agreement / UE Marketplace Standard License \\
    Rural Australia & Andrew Svanberg Hamilton & Rural environment and vegetation assets & Epic Content License Agreement / UE Marketplace Standard License \\
    MAE Beech Forest & Maelstrom Library & Forest and vegetation assets & Epic Content License Agreement / UE Marketplace Standard License \\
    Scanned 3D People Pack & Renderpeople & Scanned human character assets & Epic Content License Agreement / UE Marketplace Standard License \\
    Vehicle Variety Pack & Switchboard Studios & Vehicle assets & Epic Content License Agreement / UE Marketplace Standard License \\
    Mechanic Girl & IdaFaber & Human character asset & Epic Content License Agreement / UE Marketplace Standard License \\
    Cross-View Urban Traffic Dataset
    & Bhardwaj et al.
    & Source observations for the real-world pilot evaluation
    & CC BY-NC-ND 4.0 \\
    \bottomrule
  \end{tabular}}
\end{table*}

This work also introduces new benchmark assets, including rendered aerial--ground observations, question-answer pairs, task annotations, metadata, evaluation scripts, and documentation. We will release the benchmark data, annotations, metadata, evaluation code, and documentation the CC BY-NC 4.0 license, and the evaluation code and scripts under the MIT License. The package will include license files, attribution information, usage guidelines, and documentation of the data schema, task definitions, evaluation protocol, and metric computation. We do not redistribute Unreal Engine source code, AirSim source modifications, or third-party UE/Fab/Marketplace source assets, which remain governed by their original licenses and terms.

\subsection{Release Plan}
\label{app:release-plan}

We plan to release the AeroGround benchmark data, annotations, evaluation code, and documentation for research use. The release will include license information, attribution, usage guidelines, and scripts for reproducing the reported results. If any third-party assets cannot be redistributed due to license restrictions, we will provide instructions for obtaining the original resources and reconstructing the corresponding benchmark files where possible.

The real-world pilot set introduced in Appendix~\ref{app:real-world-pilot}, including its constructed questions and source images, is used only for the supplementary evaluation reported in this paper and is not included in the planned AeroGround release. The original real-world observations remain subject to the license and distribution terms of the source dataset.

\subsection{Responsible Use and Broader Impacts}
\label{app:responsible-use-broader-impacts}

AeroGround is designed for evaluating aerial--ground spatial understanding and may support research in navigation, embodied AI, robotics, remote sensing, and geospatial reasoning. At the same time, aerial--ground reasoning can be relevant to sensitive applications such as surveillance, tracking, and privacy-invasive monitoring.

To reduce misuse risk, we plan to release AeroGround for research and evaluation purposes and discourage applications that enable unauthorized monitoring, identification, tracking, or harmful surveillance. The benchmark is not a deployed decision-making system and is not intended for direct operational use. We also do not release tools for identifying individuals or inferring sensitive personal attributes.

\subsection{Human Subjects and LLM/VLM Usage}
\label{app:human-llm-vlm}

This work does not collect or release private or sensitive personal
information. Human evaluation is limited to benchmark question
answering under controlled instructions, and no personal attributes
are inferred or released. The four evaluators were research colleagues.
All evaluators were informed of the purpose, procedure, and intended
research use of the evaluation and provided informed consent before
participation. Participation was voluntary, and no financial
compensation was provided.

Evaluators answer questions through a unified interface that presents
the task instruction, candidate answers, paired aerial and ground
observations, and an answer panel. Hidden reference information is not
shown during evaluation.

\begin{figure}[!t]
  \centering
  \includegraphics[
    width=\linewidth,
    height=0.55\textheight,
    keepaspectratio
  ]{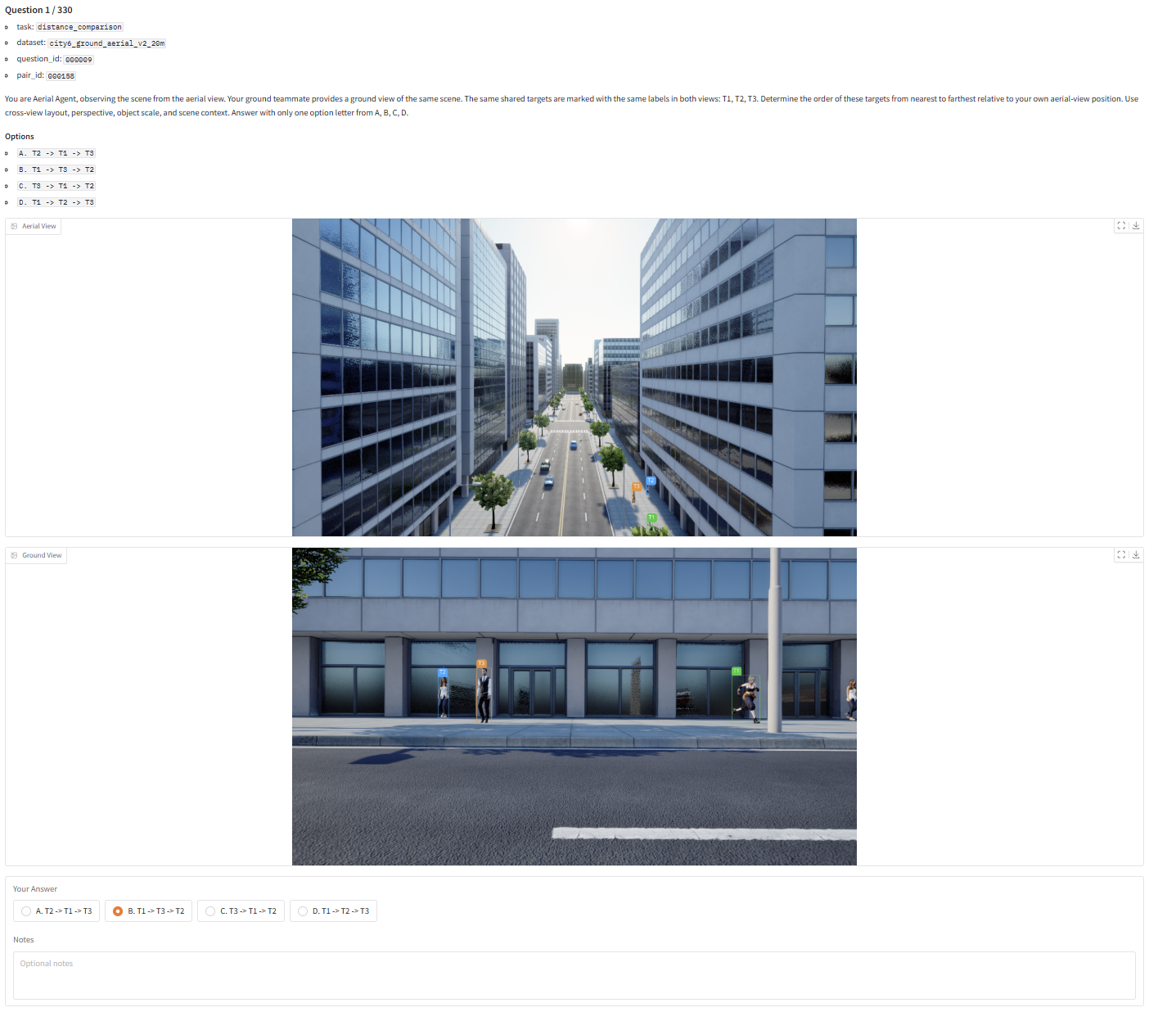}
  \caption{
  Example interface used for human evaluation. Participants are shown
  the task instruction, candidate answers, paired aerial and ground-view
  observations, and an answer panel. Hidden reference information is not
  displayed during evaluation.
  }
  \label{fig:human-eval-interface}
\end{figure}

LLMs and VLMs are used as evaluated models in the benchmark. They are
not used as a replacement for the evaluation protocol, and all reported
model results are computed using the procedure described in the paper.
\clearpage

\end{document}